\documentclass{article} 
\usepackage{iclr2027_conference,times}

\usepackage{amsmath,amsfonts,bm}

\def\eqref#1{equation~\ref{#1}}

\def\1{\bm{1}}

\DeclareMathAlphabet{\mathsfit}{\encodingdefault}{\sfdefault}{m}{sl}
\SetMathAlphabet{\mathsfit}{bold}{\encodingdefault}{\sfdefault}{bx}{n}

\DeclareMathOperator*{\argmax}{arg\,max}

\usepackage{amsmath,amssymb,amsthm,mathtools}
\usepackage{wrapfig}
\usepackage{booktabs,graphicx,xcolor,multirow}
\usepackage{algorithm,algpseudocode}
\usepackage{natbib}
\usepackage{url}
\usepackage[colorlinks,linkcolor=blue,citecolor=blue]{hyperref}

\newtheorem{proposition}{Proposition}
\newtheorem{corollary}{Corollary}
\newtheorem{lemma}{Lemma}
\newtheorem{definition}{Definition}

\newcommand{\draft}{s}
\newcommand{\cond}{c}
\newcommand{\gold}{x}

\newcommand{\pth}{p_\theta}
\newcommand{\onehot}[1]{\mathbb{1}[#1]}
\newcommand{\Cat}{\mathrm{Cat}}

\usepackage{hyperref}
\usepackage{url}
\usepackage{caption}

\title{Breaking Diffusion:\\Parallel Reasoning with Draft Refinement}

\author{Jaesang Yu$^{1}$\thanks{Correspondence to jaesangyu@kaist.ac.kr. Work done during an internship at Microsoft Research Asia.}, \; Seoyeon Byun$^{1}$, \; Yun Wang$^{2}$, \; Yan Lu$^{2}$, \; Juho Kim$^{1,3}$ \\
$^{1}$KAIST \quad $^{2}$Microsoft Research Asia \quad $^{3}$SkillBench
}

\iclrfinalcopy 

\begin{document}
\maketitle
\begin{abstract}
Drafting an answer allows a reasoning model to reconsider each decision in the context of the others. Yet repeated diffusion denoising does not ensure that individually plausible token updates form a better answer together. We introduce \textbf{Draft Refinement Language Models (DRLMs)}, which use a single denoiser to propose parallel revisions and evaluate a candidate draft. We train the denoiser to reconstruct every token from uniformly corrupted drafts, teaching both preservation and repair. Its predictions specify revisions directly, without a diffusion reverse transition. Evaluation requires a different use of these predictions: confidence in an existing token reflects both contextual support and evidence from observing the token itself. Using the known corruption process, we correct for this self-token evidence to estimate how well each token is supported by the prompt and the remaining draft. Aggregating these estimates yields a \emph{cavity energy} computed in one forward pass, without an additional scoring model. Each round proposes a revision, reads the completed candidate in its revised context, and retains it only if its energy is lower. Across Countdown, Sudoku-Extreme, and Maze-Hard, our method outperforms autoregressive and discrete diffusion baselines. With 6M parameters, DRLM reaches 44.4\% on Countdown-5 and 93.9\% on Sudoku-Extreme. The results establish complete-draft parallel refinement as an effective alternative to reverse diffusion for complex reasoning.
\end{abstract}
\section{Introduction}
\label{sec:intro}

A draft can make its own mistakes visible. Consider rereading a draft argument: its conclusion may conflict with an earlier claim, prompting us to reconsider a step that seemed plausible on its own. The draft is not merely a record of thought, but something to reason with--its parts can be inspected, compared, and revised \citep{f508427a-e4c0-3d6a-8abf-03a5d21ec6c4, 1894703.1894709}. This suggests a strategy for machine reasoning: construct a complete but provisional answer early, then use the draft itself as context for refinement. A complete draft makes the whole answer available as context, allowing earlier decisions to be reconsidered in light of later ones.

Standard autoregressive models \citep{radford2019language,brown2020language, achiam2023gpt,touvron2023llama} commits earlier tokens before a complete answer is available \citep{yao2023tree,dziri2023faith}. Discrete diffusion supports parallel prediction, but standard ancestral sampling constrains token updates through schedule-dependent reverse transitions \citep{austin2021structured,lou2024discrete,sahoo2024simple,nie2025large,ye2025dream}. Non-autoregressive refinement allows complete drafts to be revised directly \citep{lee-etal-2018-deterministic, savinov2021step}, yet individually plausible token updates do not necessarily produce a more coherent answer. The challenge is not merely to revise, but to assess its revisions. Can the same model that proposes parallel revisions also estimate the resulting draft’s contextual support in a single forward pass?

We introduce Draft Refinement Language Models (DRLMs), which use a single denoiser to propose and evaluate revisions of complete answers. We train the model to reconstruct clean sequences from drafts corrupted by uniform token replacement, supervising every position. Because incorrect tokens are unmarked, the model must learn both what to preserve and what to repair. At inference, initialized from random tokens, its clean-token predictions directly generate a revised draft: all positions are reconsidered in parallel using the same input context, without a diffusion reverse transition.

Using the same denoiser for evaluation requires separating contextual support from evidence supplied by the observed token itself. For example, in a draft such as $2+3=6$, the denoiser may favor the observed $6$ because, under the corruption channel, observing a token is far more likely when it is correct than when it is any particular alternative--even when its prediction from the remaining context, $2+3=\underline{\hspace{0.8em}}$, favors $5$. Our correction removes this self-token advantage, estimating how well the context supports the draft’s token rather than rewarding its mere presence. For a Bayes-optimal denoiser, we show that dividing out the known self-token likelihood and renormalizing recovers the token's cavity conditional \citep{mezard1987spin}: its distribution given the prompt and all other noisy draft tokens. Applying this correction to the trained model yields estimates for every position from a single full-draft read. Summing the negative log-probabilities assigned to the draft's tokens defines its \emph{cavity energy}, a learned, noise-smoothed pseudo-likelihood score \citep{besag1975}.

DRLM uses this score in a \textit{propose–read–retain} generative loop. Each round partially re-noises a copy of the retained draft to prevent a draft converging to a local optimum while preserving some of its context, then constructs a candidate through direct parallel token predictions. A fresh forward pass through the same model then scores the candidate at a fixed read level, evaluating it in its revised context rather than reusing the predictions that produced it. The candidate replaces the retained draft only if its cavity energy is lower. All proposal and scoring passes count toward a fixed inference budget, after which the decoder returns the lowest-energy draft. Our ablations show the benefits of retaining the best-scoring draft, refining it rather than restarting independently, and correcting the evaluation score for self-token evidence.

Our contributions are as follows:
\begin{itemize}

    \item \textbf{One-pass contextual draft evaluation.}
    We derive a cavity decomposition of the full clean-token posterior and use it to construct a contextual draft score from one denoiser evaluation. The construction requires neither an architectural change nor separate evaluations with individual positions removed.

    \item \textbf{Energy-controlled parallel refinement.} 
    We introduce a \textit{propose–read–retain} generative loop in which a single denoiser refines a complete draft. Starting from randomly initialized tokens, each round proposes a candidate via parallel token updates to a partially re-noised retained draft and accepts the candidate only if the denoiser assigns it lower cavity energy.

    \item \textbf{Improved reasoning performance.}
    With 6M parameters, DRLM outperforms the evaluated autoregressive and discrete diffusion baselines on Countdown, Sudoku-Extreme, and Maze-Hard. It reaches 87.6\% on Countdown-4, 44.4\% on Countdown-5, 93.9\% on Sudoku-Extreme, and 73.0\% on Maze-Hard, exceeding the strongest diffusion baseline on each task by $+14.3$, $+34.1$, $+28.4$, and $+33.7$ percentage points.
    
\end{itemize}

\section{Background}
\label{sec:background}

We consider conditional generation of a sequence $\gold=(\gold^1,\dots,\gold^L)$ over a vocabulary of size $N$ given a condition $\cond$. Interpolating discrete diffusion corrupts each position independently toward a prior $\pi$ \citep{austin2021structured,sahoo2024simple, shi2024simplified, pmlr-v267-von-rutte25a}. Using one-hot token representations and suppressing the position index, the forward channel is
\begin{equation}
q(\mathbf z_t\mid\gold)
=
\Cat\big(
\mathbf z_t;\,
\alpha_t\gold+(1-\alpha_t)\pi
\big),
\label{eq:forward}
\end{equation}
where the keep probability $\alpha_t$ decreases from $\alpha_0=1$ to $\alpha_1=0$. With $\pi=\mathbf m$, a special mask symbol, corruption either preserves a token or masks it. With $\pi=\mathbf u=\mathbf 1/N$, corruption instead replaces tokens uniformly over the ordinary vocabulary. Uniformly corrupted sequences are therefore complete drafts: every position contains an ordinary token, and replacements are unmarked and may coincide with the original tokens \citep{hoogeboom2021argmax,schiff2024discreteguidance,sahoo2025the}.

A denoiser $\gold_\theta^\ell(\cond,\mathbf z_t,t)$ uses the condition and the full noisy sequence to predict clean-token probabilities at each position. In the MDLM and UDLM plug-in parameterizations, these predictions are substituted into the categorical parameter of the exact reverse posterior $q(\mathbf z_s^\ell\mid\mathbf z_t^\ell,\gold^\ell)$ for $s<t$, yielding a factorized learned transition $\pth(\mathbf z_s\mid\cond,\mathbf z_t)$ \citep{sahoo2024simple,schiff2024discreteguidance}. Standard diffusion training minimizes a negative evidence lower bound (NELBO) \citep{pmlr-v37-sohl-dickstein15,ho2020denoising,kingma2021variational,campbell2022continuous}, whose diffusion term compares exact and learned reverse transitions. The denoiser is thus trained
through the transitions induced by its parameterization.

\paragraph{Ancestral Decoding and Token Revision.}
Given a schedule $1=t_T>\cdots>t_0=0$, standard ancestral decoding initializes each position independently from $\pi$ and samples successive states from the learned reverse transitions. In standard masked diffusion, the schedule determines the probability of revealing a masked position, while the model determines its revealed value. Once revealed, a token is copied unchanged by subsequent reverse steps,
even as its surrounding context changes \citep{sahoo2024simple}. Remasking samplers relax this restriction by allowing visible tokens to return to the mask state \citep{wang2025remasking,zheng2023rdm,ye2024beyond}.

Uniform-noise diffusion permits repeated token revisions \citep{schiff2024discreteguidance}, but the reverse kernel still constrains how predictions translate into changes. Under its plug-in parameterization, the probability of changing a position is bounded by a schedule-dependent ceiling, strictly below one for interior steps. Thus, in both standard ancestral decoders, clean-token predictions are mediated by a diffusion transition rather than specifying the next sequence directly. Appendix~\ref{app:diffusion-kernels} gives the precise parameterizations and derives the masked revelation probability and uniform revision bound.

\paragraph{From Draft Revision to Draft Evaluation.}
These constraints motivate refining complete drafts directly, letting clean-token predictions determine the next draft without a reverse diffusion transition. Direct revision, however, does not decide whether a revision should be kept: proposing a candidate and evaluating it are separate decisions. Evaluation needs care, because a full-posterior prediction treats each position's own token as evidence alongside the condition and the rest of the draft, so agreement with the current token does not by itself show contextual support.

\section{Draft Refinement Language Modeling}
\label{sec:method}

We introduce draft refinement language models (DRLMs), in which one denoiser both proposes and  evaluates revisions of complete sequences. Trained with all-position cross-entropy on uniformly corrupted drafts, it predicts full clean-token posteriors for direct parallel revision. Dividing out the known self-token likelihood turns the same predictions into one-pass cavity estimates of each token's contextual support. Each round re-noises the current draft, builds a complete candidate, and reads it in its revised context; the candidate replaces the incumbent only if its cavity energy is lower. After a fixed round budget, the decoder returns the lowest-energy candidate.

\subsection{Learning Complete-Draft Prediction}
\label{sec:training}
\label{sec:channel}
We train a single denoiser to reconstruct clean sequences from complete, uniformly corrupted drafts \citep{lee-etal-2018-deterministic,savinov2021step}, then use its predictions to propose parallel revisions. Given a condition $\cond$ and a clean sequence $\gold\in[N]^L$, we sample a keep probability
$\alpha\sim U(0,1)$ and independently corrupt each position through the uniform replacement channel:
\begin{equation}
\begin{aligned}
q_\alpha(\draft\mid\gold)
&=
\prod_{i=1}^{L} k_\alpha(\draft_i,\gold_i),\\
k_\alpha(u,v)
&=
\alpha\,\onehot{u=v}+\frac{1-\alpha}{N}.
\end{aligned}
\label{eq:channel}
\end{equation}
Each position either preserves its clean token or draws uniformly from the vocabulary, possibly drawing the same token. The resulting draft is complete: every position contains an ordinary token, and incorrect tokens are unmarked. This channel defines the denoiser's training inputs, not a diffusion trajectory that generation must
reverse.

\begin{figure}[t]
\centering
\includegraphics[width=\textwidth]{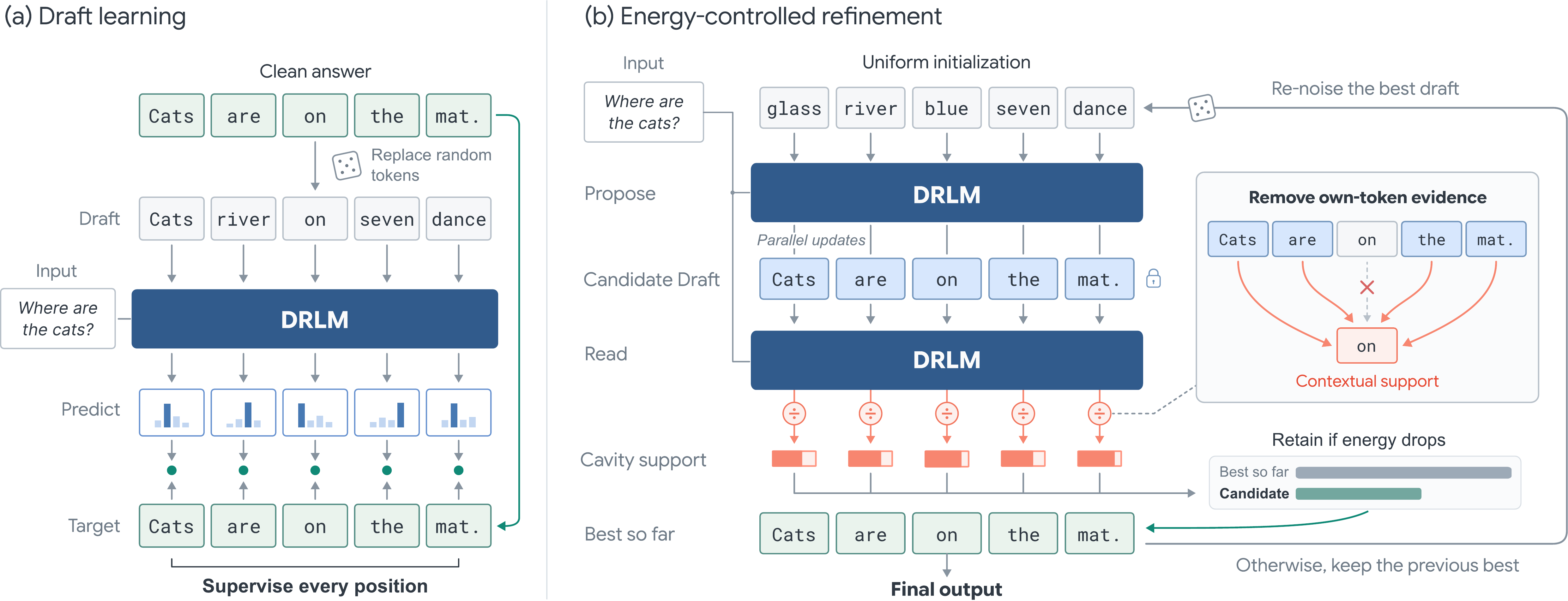}
\vspace{-10pt}
\caption{\textbf{Overview of Draft Refinement Language Models (DRLMs)}. (a) All-position supervision on uniformly corrupted drafts teaches preservation and repair. (b) Starting from uniform initialization, the decoder proposes parallel revisions and reads each completed candidate using the same denoiser, at a fixed read level and without changing its tokens. Dividing out the self-token likelihood yields cavity estimates for contextual scoring. A candidate replaces the best-so-far draft only when its cavity energy decreases; otherwise, the incumbent draft is retained. Subsequent rounds re-noise the retained draft, and decoding returns the best-energy draft after a fixed round budget.}
\label{fig:architecture}
\end{figure}

The predictor $\pth(\gold_i\mid\cond,\draft,\alpha)$ uses the condition, the entire draft, and its corruption level to predict a clean-token distribution at each position. We train it with all-position cross-entropy:
\begin{equation}
\mathcal L(\theta)
=
\mathbb E_{(\cond,\gold)}
\mathbb E_{\alpha\sim U(0,1)}
\mathbb E_{\draft\sim q_\alpha(\cdot\mid\gold)}
\left[
-\frac{1}{L}\sum_{i=1}^{L}
\log\pth(\gold_i\mid\cond,\draft,\alpha)
\right].
\label{eq:loss}
\end{equation}
Every position contributes to the loss, including those whose observed token already matches the clean token \citep{devlin2019bert,clark2020electra}. The objective therefore supervises both preservation
and repair.

With sufficient model capacity, the population minimizer recovers the clean-token posterior marginals $\Pr(\gold_i\mid\cond,\draft,\alpha)$ under the joint distribution of clean data and the corruption channel \citep{bengio2013generalized}. These are \emph{full} posterior marginals: the position's own observed token $\draft_i$ contributes evidence alongside the condition and the remaining draft.

We use these predictions directly to construct a revised draft. For a fixed condition $\cond$, define the parallel prediction map
\begin{equation}
[F_\alpha(\draft)]_i
=
\argmax_{v\in[N]}
\pth(\gold_i=v\mid\cond,\draft,\alpha),
\label{eq:prediction-map}
\end{equation}
with deterministic tie-breaking. One forward pass reconsiders all positions synchronously from the same input draft. The predicted modes determine which tokens are retained or revised, specifying the next complete draft without a diffusion reverse transition.

The training target matches this direct use of clean-token predictions. In contrast, under the plug-in reverse parameterization, the continuous-time uniform-noise NELBO targets a leave-one-out posterior, with the observed token's likelihood supplied by the reverse kernel \citep{gourevitch2026uniform}. Our cross-entropy objective instead trains the full-posterior probabilities used directly by $F_\alpha$.

\subsection{From Denoiser Predictions to Cavity Energy}
\label{sec:cavity-energy}
The prediction map specifies how to propose a revised draft, but not whether the resulting candidate should replace the current one. To evaluate complete drafts, we distinguish contextual support from the evidence supplied by each observed token itself. Full-posterior predictions incorporate both sources of evidence, as appropriate for denoising under the reconstruction training channel. For candidate evaluation, however, agreement with the input token does not by itself establish support from the condition and the remaining draft. We therefore separate the known self-token likelihood from the predictor's output.

Under the channel \eqref{eq:channel}, observing token $u=\draft_i$ supplies a likelihood ratio
\begin{equation}
\kappa_\alpha
:=
\frac{k_\alpha(u,u)}{k_\alpha(u,v)}
=
1+\frac{N\alpha}{1-\alpha},
\qquad u\neq v,\quad 0\leq\alpha<1,
\label{eq:self-likelihood-ratio}
\end{equation}
in favor of clean token $u$ over any alternative $v$. This ratio equals one at $\alpha=0$ and increases with the keep probability. For example, a $10{:}1$ channel advantage can outweigh contextual odds of $1{:}5$, leaving full-posterior odds of $2{:}1$ in favor of the observed token. A posterior preference for the draft's existing token can therefore persist even when its context favors an alternative.

Under the joint distribution of clean data and the corruption channel, define the \emph{cavity conditional} \citep{mezard1987spin} at position $i$ as
\begin{equation}
e_i^*(v\mid\cond,\draft_{-i},\alpha)
=
\Pr(\gold_i=v\mid\cond,\draft_{-i},\alpha).
\label{eq:cavity-conditional}
\end{equation}
This conditional excludes the observation $\draft_i$ while retaining the condition and all other draft tokens as noisy context. The following decomposition separates this contextual contribution from the self-token likelihood.

\begin{proposition}[Cavity decomposition]
\label{prop:cavity-decomposition}
For $0\leq\alpha<1$, the full clean-token posterior satisfies
\begin{equation}
\Pr(\gold_i=v\mid\cond,\draft,\alpha)
=
\frac{
k_\alpha(\draft_i,v)\,
e_i^*(v\mid\cond,\draft_{-i},\alpha)
}{
\sum_{u\in[N]}
k_\alpha(\draft_i,u)\,
e_i^*(u\mid\cond,\draft_{-i},\alpha)
}.
\label{eq:cavity-decomposition}
\end{equation}
\end{proposition}

The decomposition follows from Bayes' rule and independent corruption across positions. Since $k_\alpha$ is strictly positive for $\alpha<1$, we can invert the decomposition by dividing out the channel likelihood and renormalizing. Applied to the trained predictor, this gives
\begin{equation}
\widehat e_{\theta,i}(v\mid\cond,\draft,\alpha)
=
\frac{
\pth(\gold_i=v\mid\cond,\draft,\alpha)/
k_\alpha(\draft_i,v)
}{
\sum_{u\in[N]}
\pth(\gold_i=u\mid\cond,\draft,\alpha)/
k_\alpha(\draft_i,u)
}.
\label{eq:cavity-estimate}
\end{equation}
When the predictor equals the full posterior, this correction recovers $e_i^*$ exactly. All estimates are obtained from the same full-draft forward pass, without an architectural change, such as the hollow networks that output leave-one-out conditionals \citep{sun2022score,zhao2026informed}, or separate evaluations with individual positions removed \citep{salazar2020masked,goyal2021exposing}.

We use these estimates to score a complete candidate draft. Fix a read keep probability $\alpha_s\in(0,1)$, shared across candidates so that every draft is evaluated by the same scoring function. We supply the candidate $\draft$ itself as the predictor's input at $\alpha_s$; this read neither corrupts nor updates the candidate. Writing $d_i=\pth(\gold_i=\draft_i\mid\cond,\draft,\alpha_s)$, define the \emph{cavity energy} 
\begin{equation}
\begin{aligned}
E_\theta(\draft)
&=
-\sum_{i=1}^{L}
\log \widehat e_{\theta,i}
(\draft_i\mid\cond,\draft,\alpha_s)\\
&=
-\sum_{i=1}^{L}
\log
\frac{d_i}{
d_i+\kappa_{\alpha_s}(1-d_i)
}.
\end{aligned}
\label{eq:cavity-energy}
\end{equation}

We write $E_\theta$ without the fixed condition and read level. Each term measures how strongly the condition and the rest of the noisy draft support the candidate's token once that token's own channel likelihood is divided out, so lower energy means greater estimated contextual support \citep{lecun2006tutorial}. Because every term comes from the same read of the candidate, scoring a complete draft takes one forward pass and no additional scoring model. The correction is used only for evaluation: the proposal map $F_\alpha$ continues to use full-posterior predictions.

For the Bayes-optimal predictor, as $\alpha_s\uparrow1$, the cavity conditionals approach the clean-data leave-one-out conditionals, and the energy approaches negative log pseudo-likelihood \citep{besag1975}, provided every conditioning context has positive clean-data probability given $\cond$. At finite noise, the remaining draft tokens are instead treated as noisy observations. We therefore interpret $E_\theta$ as a learned, noise-smoothed pseudo-likelihood score, not a correctness certificate.

\subsection{Refinement by Proposing, Reading, and Retaining}
\label{sec:decoding}
We refine a single incumbent draft through three stages: proposing, reading, and retaining (Algorithm~\ref{alg:cavity-refinement}). Each round re-noises the incumbent and applies the parallel prediction map $F_\alpha$ \eqref{eq:prediction-map} for $J-1$ passes along an inner keep-probability schedule. The final pass reads the completed candidate at the fixed level $\alpha_s$ to compute its cavity energy, without updating its tokens. This evaluates the candidate under its revised context, rather than reusing predictions conditioned on earlier drafts. The candidate replaces the incumbent only if its energy is strictly lower. Otherwise, the incumbent and its cached energy remain unchanged, and subsequent rounds refine the retained draft. This rule maintains the \emph{best-so-far energy}. For a fixed predictor, condition, and read level, assume finite cavity energies and let
$\draft'_r$ and $\draft_r$ denote the proposal and incumbent after round $r$, respectively. Then
\begin{equation}
E_\theta(\draft_r)
=
\min_{1\leq k\leq r} E_\theta(\draft'_k),
\qquad r\geq1.
\label{eq:best-so-far-energy}
\end{equation}

\begin{wraptable}{r}{0.45\linewidth}
  \begin{minipage}{\linewidth}
    \small
    \captionsetup{font=small,skip=4pt}
    \captionof{algorithm}{Energy-controlled refinement}
    \label{alg:cavity-refinement}
    \hrule height 0.4pt
    \vspace{3pt}
  
  \begin{algorithmic}[1]
    \Require Condition $\cond$, rounds $R\geq1$,
      passes per round $J\geq2$,
      $0<\rho_e<\alpha_s<1$
    \State Initialize any $\draft\in[N]^L$;
      $E\gets+\infty$
    \For{$r=1,\ldots,R$}
      \State $\rho\gets0$ if $r=1$, else $\rho_e$
      \State $(\beta_j)_{j=1}^{J}\gets
        \operatorname{Linspace}(\rho,\alpha_s,J)$
      \State $\draft'\sim q_\rho(\cdot\mid\draft)$
        \Comment{Re-noise}
      \For{$j=1,\ldots,J-1$}
        \State $\draft'\gets F_{\beta_j}(\draft')$
          \Comment{Propose}
      \EndFor
      \State $E'\gets E_\theta(\draft')$
        \Comment{Read at $\alpha_s$; no update}
      \If{$E'<E$}
        \State $(\draft,E)\gets(\draft',E')$
          \Comment{Retain}
      \EndIf
    \EndFor
    \State \Return $\draft$
      \Comment{Best-so-far draft}
  \end{algorithmic}
    \vspace{2pt}
    \hrule height 0.4pt
  \end{minipage}
\vspace{-30pt}
\end{wraptable}
The first proposal establishes the incumbent, and every subsequent replacement strictly lowers its energy. Incumbent energy is therefore non-increasing, and the incumbent is the earliest proposal attaining the minimum.

Initialization uses the same procedure with $\rho=0$: $q_0$ supplies a uniform random canvas independently of the initial placeholder draft. Later rounds use $\rho_e$ with fresh, independent channel noise. Each round uses $J-1$ proposal passes and one candidate read, so $R$ rounds require $RJ$ model evaluations. The incumbent provides a complete best-so-far output after every round, available for return or further refinement without reevaluating its cached energy. 
\begin{figure*}[t!]
    \vspace{-10pt}
    \centering
    \includegraphics[width=\textwidth]{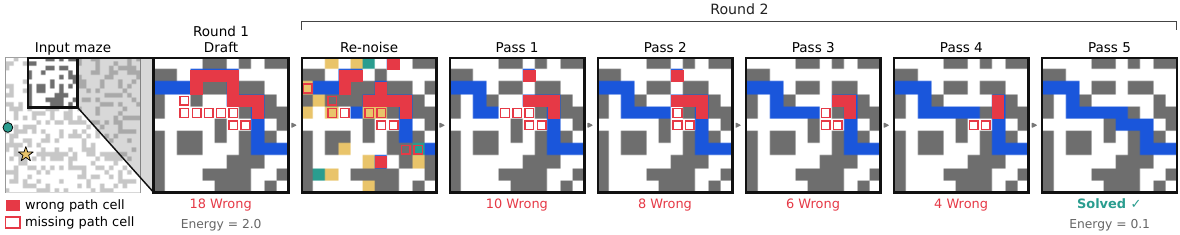}
    \vspace{-15pt}
    \caption{\textbf{Generative process on a Maze-Hard instance.}
      Given a maze with a start (circle) and a goal (star), DRLM
      generates a draft in the first round, then refines it into a valid shortest path. Each later round re-noises the draft, updates all cells in parallel, and keeps the result if its energy is lower. Panels after the input show the boxed region.}
    \label{fig:qual_mze}
    \vspace{-10pt}
\end{figure*}

\section{Experiments}
In our experiments, we evaluate our approach on three parallel reasoning tasks: Countdown, Sudoku-Extreme~\citep{wang2025hierarchicalreasoningmodel}, and Maze-Hard~\citep{wang2025hierarchicalreasoningmodel}, in which many constraint-coupled variables must be decided jointly and remain mutually consistent. These tasks require a solver to chain deductions and revise tentative assignments when they lead to contradictions. Because a single wrong position invalidates the entire solution, they test global planning rather than local pattern completion. See Appendix~\ref{app:implementation} for experimental details.

\subsection{Countdown}
Countdown is a mathematical reasoning task that generalizes the classic ``24 Game'' to arbitrary targets. Given a set of input numbers, the goal is to construct an expression that evaluates to a specified target using the operations $\{+,-,\times,\div\}$. The task tests both arithmetic accuracy and planning: a solver must identify useful intermediate
quantities and combine them into a valid solution. For example, given the numbers $\{97,38,3,17\}$ and target $14$, one valid solution is $97-38=59$, $59-17=42$, $42\div3=14$.

\paragraph{Setup.} We use the Countdown splits released by \cite{ye2024beyond}: 500k training instances with targets in 10–100, and a held-out test set of 1,000 instances. A valid solution requires the full chain to be consistent, and multiple valid answers exist per instance, so the model must commit to one complete derivation. A solution is considered correct if every equation is arithmetically valid, every operand is either an input number or an earlier result, and the final result equals the target. We study three settings of growing complexity by varying the number of inputs: CD-3, CD-4, and CD-5.

\paragraph{Baselines.}
We compare against autoregressive (AR) models at several scales:
GPT-2~\citep{radford2019language} trained from scratch with 6M, 85M, and 303M parameters, and pretrained LLaMA~\citep{touvron2023llama} with 7B and 13B parameters, fine-tuned with LoRA~\citep{hu2022lora} on the same training data. We also evaluate four paramter-matched discrete diffusion methods: the mask-absorbing models
MGDM~\citep{ye2024beyond} and LDDM~\citep{jo2026loopholing}, and the
uniform-noise models SUNDAE~\citep{savinov2021step} and UDLM~\citep{schiff2024discreteguidance}. 

\begin{wraptable}{r}{0.5\textwidth}
\vspace{-0pt}
  \centering
  \small
  \setlength{\tabcolsep}{3pt}
  \renewcommand{\arraystretch}{1.05}
  \caption{
    Results on Countdown (CD). CD-$k$ uses $k$ input numbers.
    Best results are in bold.
  }
  \vspace{-5pt}
  \label{tab:countdown}
  \begin{tabular*}{\linewidth}{@{\extracolsep{\fill}}lrrrr@{}}
    \toprule
    Model & Params & CD-3 & CD-4 & CD-5 \\
    \midrule
    \multicolumn{5}{@{}l}{\textit{Autoregressive}} \\
    GPT-2 Scratch &   6M & 94.1 & 31.9 &  4.3 \\
                  &  85M & 95.9 & 45.8 &  5.1 \\
                  & 303M & 96.4 & 41.3 &  4.5 \\
    LLaMA SFT     &   7B & 95.7 & 41.1 &  6.7 \\
                  &  13B & 96.5 & 51.1 &  7.4 \\
    \midrule
    \multicolumn{5}{@{}l}{\textit{Discrete diffusion}} \\
    SUNDAE        &   6M & 98.3 & 56.2 & 6.0 \\
    UDLM          &   6M & 98.7 & 73.3 & 6.7 \\
    MGDM          &   6M & 97.7 & 43.7 &  5.8 \\
    LDDM          &   6M & 97.6 & 56.3 & 10.3 \\
   
    \midrule
    
    \textbf{DRLM (ours)}
                  &   6M
                  & \textbf{99.5}
                  & \textbf{87.6}
                  & \textbf{44.4} \\
    \bottomrule
  \end{tabular*}
  \vspace{-15pt}
\end{wraptable}

\paragraph{Results.}
Table~\ref{tab:countdown} reports Pass@1, the fraction of problems solved by a single generated candidate. DRLM achieves the highest score across all three settings. On CD-3, it reaches near-perfect 99.5\%, exceeding UDLM by $+0.8$ percentage points. The gains are substantially larger on CD-4: DRLM achieves 87.6\%, outperforming LLaMA-13B SFT by $+36.5$ points and UDLM by $+14.3$ points. It also exceeds LDDM by $+31.3$ points.

On the more challenging CD-5 task, DRLM reaches 44.4\% accuracy, compared with 7.4\% for LLaMA-13B and 10.3\% for LDDM. These are gains of 37.0 and 34.1 percentage points, corresponding to $6.0\times$ and $4.3\times$ their respective success rates. These results support complete-draft refinement as an effective approach to parallel arithmetic reasoning.

\subsection{Sudoku and Maze-Hard}
\label{sec:exp-sudoku}
\paragraph{Sudoku-Extreme.}
Sudoku-Extreme~\citep{wang2025hierarchicalreasoningmodel} requires completing a $9\times9$ grid so that each row, column, and $3\times3$ box contains the digits $1$--$9$ exactly once. To assess how well generative models learn from data, we train on the full split of $3.8$M puzzles rather than the small-sample ($1$K) setting \citep{wang2025hierarchicalreasoningmodel,jolicoeurmartineau2025morerecursivereasoningtiny}, where performance on this unique-solution task may depend largely on architectural inductive biases. Accuracy measures the fraction of puzzles for which all $81$ predicted cells match the unique solution.

\paragraph{Maze-Hard.}
Maze-Hard~\citep{wang2025hierarchicalreasoningmodel} requires marking a shortest 4-connected path from a designated start $S$ to a goal $G$ in a $30\times30$ grid. We use $1{,}000$ training and $1{,}000$ test mazes, augmenting the training examples with the eight symmetries of the square. Unlike Sudoku-Extreme, every test maze admits multiple shortest paths. The input maze and its path-marked output are each flattened in row-major order into $900$ tokens, yielding $1{,}800$ tokens in total. A prediction is correct only if it preserves all walls and endpoints and its marked cells, together with $S$ and $G$, form a single path of optimal length, verified by breadth-first search.

\paragraph{Baselines.}
For both tasks, we compare our method with autoregressive (AR) and discrete diffusion models: LLaMA~\citep{touvron2023llama} 13B LoRA fine-tuned on the same training data and three parameter-matched diffusion models: UDLM~\citep{schiff2024discreteguidance}, MGDM~\citep{ye2024beyond}, \begin{wraptable}{r}{0.41\textwidth}
\vspace{-0pt}
  \centering
  \small
  \setlength{\tabcolsep}{3pt}
  \renewcommand{\arraystretch}{1.08}
  \caption{
    Results on Sudoku-Extreme and Maze-Hard. Best results are in bold.
  }
  \vspace{-5pt}
  \label{tab:sudoku-maze}
  \begin{tabular*}{\linewidth}{
    @{\extracolsep{\fill}}lrr@{}
  }
    \toprule
    Model
      & \shortstack[r]{Sudoku-\\Extreme}
      & \shortstack[r]{Maze-Hard} \\

    \midrule
    \multicolumn{3}{@{}l}{\textit{Autoregressive}} \\
    LLaMA SFT (13B)
      & 1.6
      & 1.5 \\
    \midrule
    \multicolumn{3}{@{}l}{\textit{Discrete diffusion}} \\
    UDLM (6M)
      & 65.5
      & 0.0 \\
    MGDM (6M)
      & 12.3
      & 10.4 \\
    LDDM (6M)
      & 24.1
      & 39.3 \\
    \midrule
    \multicolumn{3}{@{}l}{
      \textit{Specialized recurrent models}
    } \\
    HRM (27M)
      & $55.0^{\dagger}$
      & 74.5 \\
    TRM (7M)
      & $74.7^{\dagger}$
      & \textbf{85.3} \\
    \midrule
    \textbf{DRLM (6M, ours)}
      & \textbf{93.9}
      & 73.0 \\
    \bottomrule
  \end{tabular*}

  \vspace{3pt}
  \parbox{\linewidth}{
    \footnotesize\raggedright
    $\dagger$: Small-sample (1K) training and evaluation from ~\cite{jolicoeurmartineau2025morerecursivereasoningtiny}.
  }
  \vspace{-40pt}
\end{wraptable}and LDDM~\citep{jo2026loopholing}. These models use the same training splits and evaluation criteria as DRLM.

\paragraph{Results.} Table~\ref{tab:sudoku-maze} shows that our method outperforms all autoregressive and discrete diffusion baselines on both tasks. On Sudoku-Extreme, DRLM achieves $93.9\%$ exact-solution accuracy, a $+28.4$ percentage point improvement over the strongest diffusion baseline, UDLM ($65.5\%$), and a $+69.8$ point improvement over LDDM ($24.1\%$). The advantage is larger on Maze-Hard: DRLM reaches $73.0\%$ optimal-path accuracy, $+33.7$ points above LDDM ($39.3\%$). Notably, UDLM struggles on Maze-Hard. We hypothesize that this gap from the NELBO's weighting, which concentrates training signal on nearly clean canvases and thus provides little supervision for sampling $1{,}800$ tokens at once from random noise. Our results demonstrate that enabling complete-draft refinement on a standard transformer backbone unlocks substantive gains in parallel reasoning.

\section{Analysis}
\begin{figure*}[t!]
    \centering
    \includegraphics[width=\textwidth]{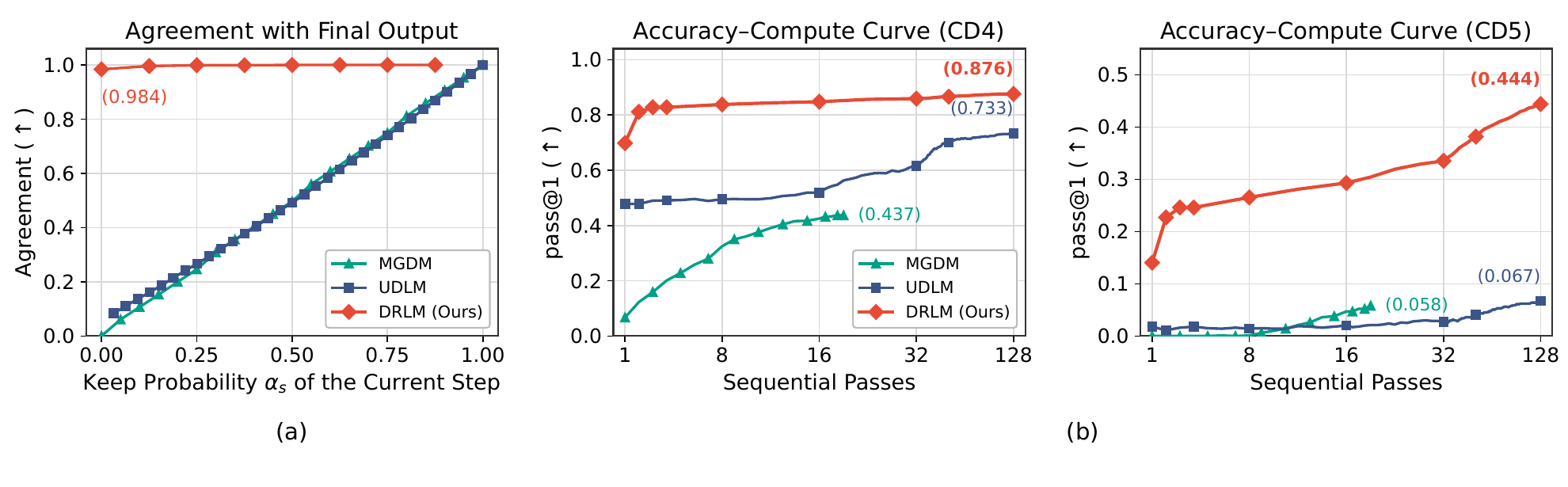}
    \vspace{-25pt}
    \caption{\textbf{Analysis.} \textbf{(a)} Agreement between each intermediate state and the decoder's own final output versus the current keep probability $\alpha_s$. DRLM reaches $98.4\%$ agreement after one round, while diffusion trajectories closely track their noise schedules. \textbf{(b)} Pass@1 of the current output after each sequential pass on Countdown-4 and Countdown-5, up to each method's default budget. DRLM exceeds both baselines' final accuracy after two passes on CD-4 and after one on CD-5.}
    \label{fig:ablation}
    \vspace{-5pt}
\end{figure*}

\paragraph{Model-bounded versus schedule-bounded generation.} Figure~\ref{fig:ablation}(a) compares decoding trajectories on held-out Countdown-4 problems using identical backbone trained on the same data. At each step, we measure the fraction of positions agreeing with the decoder's own final output. MGDM and UDLM closely track the schedule-dependent profiles $\alpha_s$ and $\alpha_s+(1-\alpha_s)/N$, respectively: agreement increases with the prescribed keep probability rather than saturating early. In contrast, DRLM already agrees with its final output at $98.4\%$ of positions after the first round, while every position remains eligible for revision. This shows that trust schedule conditions predictions rather than imposing a reverse transition: the model directly determines which tokens to retain or replace.

\paragraph{Accuracy--Compute curve.}
Figure~\ref{fig:ablation}(b) tracks the Pass@1 of each decoder's current output after every model call on Countdown-4 and Countdown-5, up to each method's default budget. On CD-4, DRLM reaches $69.8\%$ after one call and $81.1\%$ after two, already above the final accuracy of UDLM ($73.3\%$) and MGDM ($43.7\%$), and later rounds raise it to $87.6\%$. UDLM instead stays near $50\%$ for its first $16$ steps and reaches its final accuracy only at the end of its schedule. On CD-5 the gap widens: DRLM's first call ($14.0\%$) already exceeds the final accuracy of UDLM ($6.7\%$) and MGDM ($5.8\%$), and refinement keeps improving through all $32$ rounds, to $44.4\%$. DRLM thus yields usable answers from the first call and converts further calls into accuracy, most visibly on the harder task, rather than relying on a schedule to complete a solution.

\begin{table}[h]
\vspace{10pt}
\vspace{-\baselineskip}
\centering
\small
\setlength{\tabcolsep}{3pt}
\caption{Objective--Decoder ablation (Pass@1).}
\vspace{-5pt}
\label{tab:ablation}
\begin{tabular}{lcc}
\toprule
 & Reverse-diffusion & Refinement \\
Training objective & sampler (128) & decoder (128) \\
\midrule
NELBO (UDLM)          & 6.9  & 36.6 \\
All-position CE (DRLM) & 22.5 & \textbf{44.4} \\
\bottomrule
\end{tabular}
\vspace{-0.5\baselineskip}
\end{table}

\textbf{Objective and decoder contribute separately.} Table~\ref{tab:ablation} crosses the training objective and decoder on Countdown-5 at a matched budget of $128$ model evaluations, accounting for their different prediction targets. Under our refinement decoder, the UDLM model's leave-one-out predictions are converted to full posteriors. DRLM under the ancestral sampler uses the inverse conversion before applying the plug-in reverse kernel. Replacing only the decoder raises accuracy from 6.9\% to 36.6\%, while replacing only the objective gives 22.5\%. Combining both reaches 44.4\%: decoder gains appear on both networks ($+29.7$ and $+21.9$ percentage points), and objective gains under both decoders ($+15.6$ and $+7.8$); all four effects exclude zero in paired bootstrap intervals (Appendix~\ref{app:diag-objective-decoder}). Thus, both the training objective and direct refinement
contribute to the observed improvement, and their benefits
persist after accounting for the different prediction targets.

\begin{table}[h]
\vspace{-0pt}
\centering
\small
\setlength{\tabcolsep}{4pt}
\caption{Retention ablation (Pass@1).}
\vspace{-5pt}
\label{tab:retention}
\begin{tabular}{lcccc}
\toprule
Decoder & CD-4 & CD-5 & Sudoku-Extreme & Maze-Hard \\
\midrule
Accept every candidate & 82.9 & 24.0 & 68.6 & 70.3 \\
Restart from noise each round & 85.3 & 30.6 & 77.0 & 66.2 \\
\textbf{DRLM (energy-controlled retention)} & \textbf{87.6} & \textbf{44.4} & \textbf{93.9} &\textbf{73.0} \\
\bottomrule
\end{tabular}
\vspace{-10pt}
\end{table}

\paragraph{Energy-controlled retention drives refinement.}
Table~\ref{tab:retention} isolates the read-and-retain step at the same budget. Accepting every completed candidate, so that the read plays no role, leaves accuracy near its first-round level on Countdown and Sudoku-Extreme: Pass@1 falls from 87.6\% to 82.9\% on CD-4, from 44.4\% to 24.0\% on CD-5, and from 93.9\% to 68.6\% on Sudoku-Extreme. Restarting every round from a uniform canvas and keeping the lowest-energy draft recovers only part of the gain on every task (85.3\%, 30.6\%, 77.0\%, and 66.2\%), and drops to 66.2\% on Maze-Hard. On Maze-Hard, re-noising and denoising again rarely degrades a draft, while refining the retained draft remains essential. This shows that the gains come from refining a retained draft under energy control, rather than from selecting among independent drafts.
\section{Related Work}

\paragraph{Discrete Diffusion.}
Discrete diffusion learns reverse transitions for categorical corruption processes \citep{austin2021structured,hoogeboom2021argmax,campbell2022continuous,lou2024discrete}, including masking \citep{sahoo2024simple,shi2024simplified,ou2025your} and uniform replacement \citep{schiff2024discreteguidance,sahoo2025the}, and has been scaled to billion-parameter language models \citep{nie2025large,ye2025dream, ye2026sumi}. MGDM adapts masked diffusion to parallel reasoning \citep{ye2024beyond}, adaptive unmasking orders improve masked diffusion on Sudoku \citep{kim2025train}, and LDDM propagates continuous information alongside sampled tokens \citep{jo2026loopholing}. Remasking and correction methods enable revision through remasking \citep{zheng2023rdm,wang2025remasking},
mixed mask--uniform corruption \citep{pmlr-v267-von-rutte25a}, explicit training on incorrect visible tokens \citep{zhang2025correctivediffusionlanguagemodels,schiff2026learn}, learned per-token quality scores that select tokens to remask \citep{kim2025fine}, or corrector steps that resample tokens from leave-one-out conditionals \citep{zhao2026informed}. These methods retain generation based on diffusion transitions, remasking, or interleaved correction steps. Our method instead uses uniform corruption only to construct training drafts without sampling a diffusion reverse transition, and it uses the channel likelihood to score complete candidates rather than to resample individual tokens.

\paragraph{Non-Autoregressive Iterative Refinement.}
Complete-sequence refinement predates discrete diffusion: deliberation networks polish a first-pass draft \citep{xia2017deliberation}, and non-autoregressive translation refines drafts by iterative denoising \citep{lee-etal-2018-deterministic} or by insertion and deletion edits \citep{NIPS2019_9297}. Mask-Predict \citep{ghazvininejad-etal-2019-mask} and DisCo \citep{pmlr-v119-kasai20a} develop confidence- and context-dependent parallel updates. SUNDAE \citep{savinov2021step} learns clean-sequence reconstruction from uniformly corrupted inputs with its self-generated trajectories, and alternating corruption and denoising also defines generative Markov chains \citep{bengio2013generalized}. Unlike these methods, which keep every revision, DRLM treats each revision as a proposal: it keeps the revised draft only if the same model assigns it lower energy. Recent structured reasoners scale test-time computation through weight-tied recurrence: HRM uses hierarchical recurrent states \citep{wang2025hierarchicalreasoningmodel}, while TRM
\citep{jolicoeurmartineau2025morerecursivereasoningtiny} and PTRM \citep{sghaier2026probabilistic} update both answer and latent states. DRLM instead uses the complete discrete draft itself as the recurrent state, making each round's candidate directly interpretable.

\section{Conclusion}
\label{sec:conclusion}
We introduced Draft Refinement Language Models (DRLMs), which use a single denoiser to propose and evaluate parallel revisions of complete drafts. All-position cross-entropy training on uniformly corrupted sequences teaches preservation and repair, enabling direct updates without a diffusion reverse transition. Correcting predictions for the known self-token likelihood yields cavity estimates for one-pass contextual evaluation, supporting a \textit{propose–read–retain} generative loop that returns the lowest-energy evaluated draft within a fixed budget. Across Countdown, Sudoku-Extreme, and Maze-Hard, our method outperforms the evaluated autoregressive and discrete diffusion baselines. These results demonstrate the effectiveness of combining parallel revision with whole-draft evaluation for complex parallel reasoning.
\bibliography{iclr2027_conference}
\bibliographystyle{iclr2027_conference}

\appendix
\section{Appendix}

\section{Use of Large Language Models}
During the preparation of this manuscript, we utilized a large language model to improve grammar, clarity, and overall readability.

\section{Diffusion Kernels and Token-Revision Constraints}
\label{app:diffusion-kernels}

This appendix makes precise the statements about standard ancestral decoding in Section~\ref{sec:background}: the plug-in reverse kernels of MDLM and UDLM, the schedule-controlled revelation of masked diffusion, and the revision ceiling of uniform-noise diffusion. Throughout, $\tau<t$ are two diffusion times with keep probabilities $b=\alpha_\tau>a=\alpha_t\geq0$. These times are unrelated to the read level $\alpha_s$ of Section~\ref{sec:cavity-energy}. We fix one position, the condition, and the current state $\mathbf z_t$, and write $u$ for the current token at that position and $v$ for a possible next token.

\subsection{Plug-in Reverse Kernels}
\label{app:plugin-kernel}

Under the forward channel \eqref{eq:forward}, the transition between the two times is
\begin{equation}
q(\mathbf z_t=u\mid\mathbf z_\tau=v)=\gamma\onehot{u=v}+(1-\gamma)\pi_u,
\qquad
\gamma=\frac{a}{b}\in[0,1),
\label{eq:app-forward-bridge}
\end{equation}
which composes with $q(\mathbf z_\tau\mid\gold)$ to give $q(\mathbf z_t\mid\gold)$. For a clean token with one-hot vector $\mathbf e_{\gold}$, Bayes' rule gives the exact reverse posterior
\begin{equation}
q(\mathbf z_\tau=v\mid\mathbf z_t=u,\gold)
=
\frac{[\gamma\onehot{v=u}+(1-\gamma)\pi_u]\,[b(\mathbf e_{\gold})_v+(1-b)\pi_v]}
{a(\mathbf e_{\gold})_u+(1-a)\pi_u}.
\label{eq:app-exact-posterior}
\end{equation}

\begin{definition}[Plug-in kernel]
\label{def:app-plugin-kernel}
For a probability vector $f$, the denoiser's prediction at the position, the plug-in kernel substitutes $f$ for $\mathbf e_{\gold}$ in \eqref{eq:app-exact-posterior}:
\begin{equation}
B_f^\pi(v\mid u)
=
\frac{[\gamma\onehot{v=u}+(1-\gamma)\pi_u]\,[bf_v+(1-b)\pi_v]}
{af_u+(1-a)\pi_u},
\qquad
af_u+(1-a)\pi_u>0.
\label{eq:app-plugin-kernel}
\end{equation}
Applied independently at every position, it is the factorized learned transition $\pth(\mathbf z_\tau\mid\cond,\mathbf z_t)$ of MDLM and UDLM \citep{sahoo2024simple,schiff2024discreteguidance}.
\end{definition}

\begin{lemma}
\label{lem:app-plugin-normalized}
(i) $\sum_vB_f^\pi(v\mid u)=1$. (ii) $B_{\mathbf e_{\gold}}^\pi$ is the exact posterior \eqref{eq:app-exact-posterior}. (iii) In general, $B_f^\pi$ differs from the average $\sum_{\gold}f_{\gold}\,q(\mathbf z_\tau=\cdot\mid\mathbf z_t=u,\gold)$ of exact posteriors.
\end{lemma}
\begin{proof}
(i) Summing the numerator over $v$ gives $\gamma[bf_u+(1-b)\pi_u]+(1-\gamma)\pi_u=af_u+(1-a)\pi_u$, because $\gamma b=a$. (ii) is immediate. (iii) Take the uniform prior and the clean endpoint $b=1$. Each exact posterior is then the point mass at $\gold$, so the average equals $f$, whereas \eqref{eq:app-plugin-kernel} gives $B_f^{\mathbf u}(v\mid u)\propto k_a(u,v)\,f_v$ with $k_a$ from \eqref{eq:channel}: the plug-in kernel reweights the prediction by the likelihood of the current token.
\end{proof}

The substitution thus includes the normalizer, rather than averaging separately normalized posteriors.

\subsection{Masked Revelation and Carry-Over}
\label{app:masked-kernel}

Let $\pi=\mathbf m$ be the mask, and adopt MDLM's parameterization: the prediction assigns no probability to the mask, $f_{\mathbf m}=0$, and visible tokens are carried over \citep{sahoo2024simple}.

\begin{proposition}[Schedule-controlled revelation]
\label{prop:app-masked}
(i) At a masked position,
\begin{equation}
B_f^{\mathbf m}(v\mid\mathbf m)
=
\frac{1-b}{1-a}\onehot{v=\mathbf m}+\frac{b-a}{1-a}f_v,
\qquad
\Pr_\theta(\mathbf z_\tau\neq\mathbf m\mid\cond,\mathbf z_t)
=
\frac{\alpha_\tau-\alpha_t}{1-\alpha_t},
\label{eq:app-masked-revelation}
\end{equation}
and a revealed token is distributed as $f$. (ii) At a visible position $u\neq\mathbf m$ with $af_u>0$, $B_f^{\mathbf m}(\cdot\mid u)$ is the point mass at $u$, and MDLM's carry-over imposes this for every $f$. Hence a revealed token is never changed by later reverse steps.
\end{proposition}
\begin{proof}
(i) With $\pi=\mathbf m$ and $u=\mathbf m$, the numerator of \eqref{eq:app-plugin-kernel} is $1-b$ for $v=\mathbf m$ and $(1-\gamma)bf_v=(b-a)f_v$ otherwise, and the denominator is $1-a$. Summing over ordinary tokens gives the revelation probability. (ii) For $u\neq\mathbf m$, $\pi_u=0$, so the numerator vanishes unless $v=u$, where it equals $\gamma bf_u=af_u$, the denominator. Under the forward law, a visible token at time $t$ requires $a>0$.
\end{proof}

The revelation probability is set by the schedule, and the model determines only the revealed value. The proposition concerns this standard kernel. Remasking samplers \citep{wang2025remasking,zheng2023rdm,ye2024beyond} and confidence-ordered decoders modify it and are outside its scope.

\subsection{Uniform-Noise Revision Ceiling}
\label{app:uniform-kernel}

Let $\pi=\mathbf u=\mathbf 1/N$ over the $N$ ordinary tokens, and write $\nu=bf+(1-b)\mathbf u$ and $D=af_u+(1-a)/N>0$.

\begin{proposition}[Redraw form]
\label{prop:app-uniform-mixture}
The uniform plug-in kernel mixes staying with a redraw from $\nu$:
\begin{equation}
B_f^{\mathbf u}(\cdot\mid u)=(1-\lambda_f)\,\delta_u+\lambda_f\,\nu,
\qquad
\lambda_f=\frac{1-\gamma}{ND}=\frac{b-a}{b\,(1-a+Naf_u)}\in(0,1].
\label{eq:app-uniform-mixture}
\end{equation}
For $b<1$, every token has redraw probability $\nu_v\geq(1-b)/N>0$, so visible tokens are not absorbing.
\end{proposition}
\begin{proof}
The numerator of \eqref{eq:app-plugin-kernel} is $[\gamma\onehot{v=u}+(1-\gamma)/N]\,\nu_v$. Since $\gamma\nu_u=af_u+\gamma(1-b)/N$, we have $D-\gamma\nu_u=(1-\gamma)/N$. Hence $1-\lambda_f=\gamma\nu_u/D\in[0,1)$, and the two expressions for $\lambda_f$ agree.
\end{proof}

\begin{proposition}[Revision ceiling]
\label{prop:app-uniform-ceiling}
Since a redraw may return the current token,
\begin{equation}
\begin{aligned}
\Pr_\theta(\mathbf z_\tau\neq u\mid\cond,\mathbf z_t)
&=\lambda_f(1-\nu_u)
=\frac{(b-a)\,[1-bf_u-(1-b)/N]}{b\,(1-a+Naf_u)}\\
&\leq C_{\tau,t}\Big(1-\frac{1-b}{N}\Big)
\leq C_{\tau,t},
\qquad
C_{\tau,t}=\frac{b-a}{b(1-a)}.
\end{aligned}
\label{eq:app-uniform-ceiling}
\end{equation}
The first bound is attained exactly when $f_u=0$, strictly positive softmax outputs approach it without reaching it. For interior steps $0<a<b<1$, $1-C_{\tau,t}=a(1-b)/[b(1-a)]>0$, so the change probability is bounded strictly below one, whatever the prediction.
\end{proposition}
\begin{proof}
In \eqref{eq:app-uniform-ceiling}, the numerator decreases and the denominator does not decrease in $f_u$, so the change probability is strictly decreasing in $f_u$ and its value at $f_u=0$ is the first bound.
\end{proof}

Equivalently, $C_{\tau,t}(1-(1-b)/N)=1-\frac{1-b}{1-a}\big(\gamma+\frac{1-\gamma}{N}\big)$. The ceiling is vacuous at the boundaries: for $a=0$, $\lambda_f=1$ and the kernel redraws from $\nu$. For $b=1$, $C_{\tau,t}=1$, and $f_u=0$ changes the token with probability one. The kernel \eqref{eq:app-uniform-mixture} is the one used by UDLM's ancestral sampler \citep{schiff2024discreteguidance} and the ceiling is our derivation from it.

\section{Full-Posterior Learning and Direct Prediction}
\label{app:training-details}

Throughout, $(\cond,\gold)$ is drawn from the data, $\alpha\sim U(0,1)$, and $\draft\sim q_\alpha(\cdot\mid\gold)$ from \eqref{eq:channel}. We write $p_i^*(v\mid\cond,\draft,\alpha)=\Pr(\gold_i=v\mid\cond,\draft,\alpha)$ for the full posterior, $e_i^*$ for the cavity conditional \eqref{eq:cavity-conditional}, and $\varepsilon_\alpha=(1-\alpha)/N$, so that $k_\alpha(u,v)=\alpha\onehot{u=v}+\varepsilon_\alpha$.

\subsection{Population Optimum of the All-Position Cross-Entropy}
\label{app:ce-optimum}

\begin{proposition}
\label{prop:app-ce-minimizer}
For any predictor $f$ with outputs $f_i(\cdot\mid\cond,\draft,\alpha)$ over the model vocabulary $\mathcal V\supseteq[N]$,
\begin{equation}
\mathcal L(f)-\mathcal L(p^*)
=
\mathbb E_{\cond,\draft,\alpha}\Big[\frac1L\sum_{i=1}^{L}\mathrm{KL}\big(p_i^*\,\|\,f_i\big)\Big]
\geq0,
\label{eq:app-ce-decomposition}
\end{equation}
where $\mathcal L$ is the objective \eqref{eq:loss} and $\mathcal L(p^*)\leq\log N$. Equality holds if and only if $f_i=p_i^*$ at every position almost surely. In particular, the optimum assigns no probability outside $[N]$.
\end{proposition}
\begin{proof}
Given $(\cond,\draft,\alpha)$, the expected loss at position $i$ is $-\sum_vp_i^*(v)\log f_i(v)=H(p_i^*)+\mathrm{KL}(p_i^*\|f_i)$, with $H(p_i^*)\leq\log N$. Averaging over positions and inputs gives \eqref{eq:app-ce-decomposition}, and each KL term vanishes only when the two distributions agree.
\end{proof}

The optimum is characterized through its marginals: no independence between clean tokens is assumed, and the objective does not identify a joint distribution over sequences. A trained network approximates $p^*$ within its capacity, data, and optimization budget, whereas the proposition concerns the population optimum.

\subsection{Preservation, Repair, and Synchronous Prediction}
\label{app:preserve-repair}

Because each position either keeps its clean token or is replaced uniformly, an observed token can match the clean token in two ways:
\begin{equation}
\Pr(\draft_i=\gold_i\mid\gold_i,\alpha)=\alpha+\varepsilon_\alpha,
\qquad
\Pr(\text{kept}\mid\draft_i=\gold_i,\alpha)=1-\frac{1}{\kappa_\alpha}.
\label{eq:app-match-prob}
\end{equation}
The loss \eqref{eq:loss} supervises matching positions, teaching preservation, and mismatching positions, teaching repair; preservation is learned from this supervision rather than imposed by a copy rule. The map $F_\alpha$ \eqref{eq:prediction-map} takes the positionwise modes of $\pth$ for one common input draft, synchronously at all positions, with ties broken toward the lowest token index. It is neither an autoregressive sweep nor a maximizer of the joint probability of the sequence, and one application need not lower the cavity energy. In our implementation the mode is taken over the full model vocabulary; a proposal containing a token outside the $N$-symbol canvas alphabet cannot be scored and is never retained (Appendix~\ref{app:cavity-impl}). No such proposal occurred in any reported run.

\subsection{Training conventions}
\label{app:training-conventions}

Each training sequence draws one keep probability, shared by all its positions: $\alpha=1-U$ with $U\sim U[0,1)$. The network receives the replacement rate $1-\alpha$ through the features $\{\sin(\pi k(1-\alpha)),\cos(\pi k(1-\alpha))\}_{k=1}^{5}$, mapped by a two-layer MLP and added to every canvas embedding. Condition tokens are never corrupted. The loss covers all $L$ canvas positions, including padding and end-of-sequence positions in Countdown, with the softmax over the full model vocabulary $\mathcal V$. Corruption and scoring use the $N$-symbol canvas alphabet. Architecture, optimizer, and budgets are given in Table~\ref{tab:app-train-config}.

\paragraph{The endpoint $\alpha=0$.}
The first decoding round applies $F_0$ to a uniformly random canvas (Algorithm~\ref{alg:cavity-refinement} with $\rho=0$). Training never samples $\alpha=0$ exactly, so $F_0$ relies on continuity at the endpoint. Under the exact posterior a canvas at $\alpha=0$ carries no information about $\gold$, and the first pass would ignore it. The learned $F_0$ does not: on Countdown-4, the Pass@1 of the first pass differs across decode seeds ($69.4\%$, $69.9\%$, $70.2\%$). We make no claim that uniform initialization alone produces diverse drafts.

\section{Cavity energy derivations and implementation}
\label{app:cavity-energy}

\subsection{Proof of the Cavity Decomposition}
\label{app:cavity-proof}

\begin{lemma}
\label{lem:app-self-obs}
Under the joint law $p(\cond,\gold)\,q_\alpha(\draft\mid\gold)$, the observation $\draft_i$ is conditionally independent of $(\cond,\gold_{-i},\draft_{-i})$ given $\gold_i$ and $\alpha$, and $\Pr(\draft_i=u\mid\gold_i=v,\alpha)=k_\alpha(u,v)$.
\end{lemma}
\begin{proof}
The joint density $p(\cond,\gold)\prod_jk_\alpha(\draft_j,\gold_j)$ depends on $\draft_i$ only through the factor $k_\alpha(\draft_i,\gold_i)$, which sums to one over $\draft_i$.
\end{proof}

\begin{proof}[Proof of Proposition~\ref{prop:cavity-decomposition}]
By Bayes' rule and Lemma~\ref{lem:app-self-obs},
\[
\Pr(\gold_i=v\mid\cond,\draft,\alpha)
=
\frac{\Pr(\draft_i\mid\gold_i=v,\alpha)\,\Pr(\gold_i=v\mid\cond,\draft_{-i},\alpha)}
{\sum_u\Pr(\draft_i\mid\gold_i=u,\alpha)\,\Pr(\gold_i=u\mid\cond,\draft_{-i},\alpha)}
=
\frac{k_\alpha(\draft_i,v)\,e_i^*(v)}{\sum_uk_\alpha(\draft_i,u)\,e_i^*(u)}.
\]
For $\alpha<1$ the denominator is at least $\varepsilon_\alpha>0$.
\end{proof}

What is excluded is the observation $\draft_i$: the cavity conditional still conditions on $\cond$ and on every other noisy token $\draft_{-i}$, and $\gold_i$ remains the variable being predicted.

\subsection{Likelihood Inversion and Exact Recovery}
\label{app:cavity-inversion}

For $\alpha<1$, $k_\alpha(u,v)\geq\varepsilon_\alpha>0$, so dividing \eqref{eq:cavity-decomposition} by $k_\alpha(\draft_i,\cdot)$ and renormalizing inverts it.

\begin{corollary}[Exact recovery]
\label{cor:app-exact-recovery}
If $\pth=p^*$, the estimate \eqref{eq:cavity-estimate} equals $e_i^*(\cdot\mid\cond,\draft_{-i},\alpha)$ and therefore does not depend on $\draft_i$. At the draft's own token, with $d_i=\pth(\gold_i=\draft_i\mid\cond,\draft,\alpha)$,
\begin{equation}
\widehat e_{\theta,i}(\draft_i\mid\cond,\draft,\alpha)=\frac{d_i}{d_i+\kappa_\alpha(1-d_i)},
\label{eq:app-self-cavity}
\end{equation}
which gives the second line of \eqref{eq:cavity-energy}.
\end{corollary}
\begin{proof}
The first claim inverts Proposition~\ref{prop:cavity-decomposition}. For \eqref{eq:app-self-cavity}, $k_\alpha(\draft_i,\cdot)$ equals $\alpha+\varepsilon_\alpha$ at $\draft_i$ and $\varepsilon_\alpha$ elsewhere, so the normalizer of \eqref{eq:cavity-estimate} is $d_i/(\alpha+\varepsilon_\alpha)+(1-d_i)/\varepsilon_\alpha$. Dividing by it and using $\kappa_\alpha=(\alpha+\varepsilon_\alpha)/\varepsilon_\alpha$ gives the result.
\end{proof}

\begin{corollary}[Odds form]
\label{cor:app-odds}
For $v\neq u=\draft_i$ with $e_i^*(v)>0$,
\[
\frac{p_i^*(u)}{p_i^*(v)}=\kappa_\alpha\,\frac{e_i^*(u)}{e_i^*(v)},
\qquad
\kappa_\alpha=1+\frac{N\alpha}{1-\alpha},
\]
where $\kappa_0=1$, $\kappa_\alpha$ is strictly increasing, and $\kappa_\alpha\to\infty$ as $\alpha\uparrow1$.
\end{corollary}

The self-likelihood ratio multiplies the observed token's odds against every alternative. It is local to the position, whereas $e_i^*$ depends on $\alpha$ through the noisy context. For an approximate predictor, \eqref{eq:cavity-estimate} defines an estimate: the correction removes the channel's known self-likelihood, not arbitrary model error, and $\widehat e_{\theta,i}$ may still depend on $\draft_i$.

\subsection{Scoring Completed Candidates}
\label{app:cavity-read}

Each candidate is scored by a fresh read: the completed candidate itself is the input, at the common read level $\alpha_s$. The last proposal pass cannot serve as this read, because its predictions were conditioned on the draft that preceded the candidate rather than on the candidate. The read samples no corruption, replaces no token, and does not apply $F_{\alpha_s}$. A single forward pass returns the logits from which all $L$ cavity estimates are computed. The self-likelihood correction is applied only here: proposals use the full-posterior predictions.

\subsection{Clean-Data Pseudo-Likelihood Limit}
\label{app:pl-limit}

For the Bayes-optimal predictor and a fixed candidate $\draft$, the cavity conditional is a ratio of finite sums over clean sequences:
\begin{equation}
e_i^*(v\mid\cond,\draft_{-i},\alpha)
=
\frac{\sum_{\gold:\gold_i=v}p(\gold\mid\cond)\prod_{j\neq i}k_\alpha(\draft_j,\gold_j)}
{\sum_{\gold}p(\gold\mid\cond)\prod_{j\neq i}k_\alpha(\draft_j,\gold_j)}.
\label{eq:app-noisy-cavity}
\end{equation}

\begin{proposition}[Pseudo-likelihood limit]
\label{prop:app-pl-limit}
Assume $p(\gold_{-i}=\draft_{-i}\mid\cond)>0$ for every $i$. As $\alpha\uparrow1$,
\begin{gather*}
e_i^*(v\mid\cond,\draft_{-i},\alpha)\to p(\gold_i=v\mid\cond,\gold_{-i}=\draft_{-i}),\\
E^*_\alpha(\draft)=-\sum_{i=1}^{L}\log e_i^*(\draft_i\mid\cond,\draft_{-i},\alpha)\to-\log\mathrm{PL}(\draft\mid\cond)\in[0,\infty],
\end{gather*}
where $\mathrm{PL}(\draft\mid\cond)=\prod_ip(\gold_i=\draft_i\mid\cond,\gold_{-i}=\draft_{-i})$ is the pseudo-likelihood \citep{besag1975}. The limit is finite if and only if every conditional is positive. At $\alpha=0$, $e_i^*(v)=p(\gold_i=v\mid\cond)$.
\end{proposition}
\begin{proof}
As $\alpha\uparrow1$, $k_\alpha(u,v)\to\onehot{u=v}$, so each finite sum in \eqref{eq:app-noisy-cavity} converges to its terms with $\gold_{-i}=\draft_{-i}$: the numerator to $p(\gold_i=v,\gold_{-i}=\draft_{-i}\mid\cond)$ and the denominator to $p(\gold_{-i}=\draft_{-i}\mid\cond)>0$. The energy converges in $[0,\infty]$ by continuity of $-\log$ on $[0,1]$. At $\alpha=0$, the constant kernel $k_0\equiv1/N$ cancels.
\end{proof}

If a context has zero probability under the data, the argument does not define the clean conditional. The approach to the limit can be slow when $p(\gold_{-i}=\draft_{-i}\mid\cond)$ is small, because sequences that differ from $\draft_{-i}$ in one position enter \eqref{eq:app-noisy-cavity} with relative weight $O(1-\alpha)$. The limit concerns the exact posterior, and a learned predictor inherits it only insofar as $\pth$ approximates $p^*$ near $\alpha=1$. At the finite read level used in practice, the other draft tokens act as noisy observations, which is why we describe $E_\theta$ as a noise-smoothed pseudo-likelihood.

\subsection{Stable Logit-Space Implementation}
\label{app:cavity-impl}

Let $h_{i,v}$ be the read's logits restricted to the canvas alphabet. Since $(1-d_i)/d_i=\exp(\operatorname{logsumexp}_{v\neq\draft_i}h_{i,v}-h_{i,\draft_i})$, the energy \eqref{eq:cavity-energy} equals
\begin{equation}
E_\theta(\draft)
=\sum_{i=1}^{L}\operatorname{softplus}\!\Big(\log\kappa_{\alpha_s}+\operatorname*{logsumexp}_{v\neq\draft_i}h_{i,v}-h_{i,\draft_i}\Big)
=\sum_{i=1}^{L}\Big[\operatorname*{logsumexp}_{v}\widetilde h_{i,v}-\widetilde h_{i,\draft_i}\Big],
\label{eq:app-logit-energy}
\end{equation}
with corrected logits $\widetilde h_{i,v}=h_{i,v}-\log\kappa_{\alpha_s}\onehot{v=\draft_i}$ and $\log\kappa_{\alpha_s}=\operatorname{log1p}((N-1)\alpha_s)-\operatorname{log1p}(-\alpha_s)$. Neither form evaluates $1-d_i$, which is lost to rounding when $d_i$ is close to one. Our implementation computes the energy of a batch of candidates from one forward pass:
\begin{scriptsize}
\begin{verbatim}
logits = net(cond, cand, level=1 - alpha_s)        # bf16 forward pass, [B, L, |V|]
lp     = log_softmax(logits[..., canvas].float())  # fp32, over the N canvas symbols
own    = one_hot(canvas_index(cand), N)            # [B, L, N]
field  = lp + log_kappa * (1 - own)                # log kappa added off the own token
E      = (logsumexp(field, -1) - (lp * own).sum(-1)).sum(-1)
E[candidates holding a non-canvas token] = +inf
\end{verbatim}
\end{scriptsize}
Adding $\log\kappa_{\alpha_s}$ to every entry except the own token is equivalent to subtracting it from the own token, because each term of \eqref{eq:app-logit-energy} is invariant to a common shift of the logits. We apply no clipping and no tolerance: a candidate is retained only if its energy is strictly lower, and the incumbent's energy is cached. With $\alpha_s=7/8$, $\kappa_{\alpha_s}=1+7N$, that is, $127$ for Countdown ($N=18$), $64$ for Sudoku-Extreme ($N=9$), and $36$ for Maze-Hard ($N=5$). We checked this implementation against a float64 evaluation of \eqref{eq:cavity-estimate}, of the $d_i$-form, and of the softplus form on random logits for $\alpha_s\in\{0.5,0.875,0.98\}$. The largest discrepancy, $7\times10^{-5}$ on energies summed over $64$ positions, is at float32 rounding level.

\subsection{Boundary Cases and Interpretation}
\label{app:cavity-boundary}

At $\alpha_s=0$ the correction vanishes ($\kappa_0=1$) and, for the exact posterior, each term reduces to the condition-only marginal $p(\gold_i=\draft_i\mid\cond)$, which ignores the rest of the draft. As $\alpha_s\uparrow1$ the ideal score approaches the clean pseudo-likelihood (Proposition~\ref{prop:app-pl-limit}), but $\log\kappa_{\alpha_s}$ diverges and amplifies errors in the predicted $1-d_i$. At $\alpha_s=1$ the kernel is the identity and cannot be inverted. We use the interior value $\alpha_s=7/8$ (Appendix~\ref{app:diag-sweep} reports the sensitivity to it). The energy sums positionwise log-conditionals, so it is not a normalized sequence likelihood, and a lower energy does not certify correctness. All comparisons fix the condition and the sequence length.

\section{Refinement Algorithm and Reproducibility}
\label{app:decoding}

\subsection{Decoding Specification}
\label{app:decoding-spec}

Algorithm~\ref{alg:app-refinement-batched} expands Algorithm~\ref{alg:cavity-refinement} for a batch of conditions, with separate incumbent and candidate buffers. Each round uses the inner keep-probability schedule
\begin{equation}
\beta_j=\rho+\frac{j-1}{J-1}\,(\alpha_s-\rho),\qquad j=1,\dots,J,
\label{eq:app-inner-schedule}
\end{equation}
applying $F_{\beta_1},\dots,F_{\beta_{J-1}}$ and reading at $\beta_J=\alpha_s$ without a token update. For $J=2$, a round is one proposal pass at $\beta_1=\rho$ followed by the read. In the first round $\rho=0$, so the re-noised canvas is uniform on $[N]^L$ and independent of the placeholder (our implementation starts from an all-padding placeholder, which is never scored). Later rounds use $\rho=\rho_e$ with fresh, independent noise. Re-noising replaces a token with probability $1-\rho$, and the uniform draw may return the same token, so the expected fraction of changed positions is
\begin{equation}
\Pr(\draft'_i\neq\draft_i)=(1-\rho)\Big(1-\frac1N\Big),
\label{eq:app-renoise-change}
\end{equation}
not $1-\rho$. Our code parameterizes noise by the replacement rate $1-\beta$. The schedule is the same linear spacing expressed in that variable.

\begin{algorithm}[t]
\caption{Energy-controlled refinement, batched form of Algorithm~\ref{alg:cavity-refinement}}
\label{alg:app-refinement-batched}
\begin{algorithmic}[1]
\Require conditions $\cond^{(1)},\dots,\cond^{(B)}$; rounds $R\geq1$; passes per round $J\geq2$; $0<\rho_e<\alpha_s<1$
\State $\draft^{(b)}\gets$ placeholder, $E^{(b)}\gets+\infty$ for all $b$ \Comment{incumbent and cached energy}
\For{$r=1,\dots,R$}
  \State $\rho\gets0$ if $r=1$, else $\rho_e$; compute $\beta_1,\dots,\beta_J$ by \eqref{eq:app-inner-schedule}
  \State $\draft'^{(b)}\sim q_\rho(\cdot\mid\draft^{(b)})$ for all $b$, with fresh noise \Comment{candidate buffer}
  \For{$j=1,\dots,J-1$}
    \State $\draft'^{(b)}\gets F_{\beta_j}(\draft'^{(b)})$ for all $b$ \Comment{one batched forward pass}
  \EndFor
  \State $E'^{(b)}\gets E_\theta(\draft'^{(b)})$ at $\alpha_s$ for all $b$ \Comment{read; no token update}
  \For{each $b$ with $E'^{(b)}<E^{(b)}$}
    \State $(\draft^{(b)},E^{(b)})\gets(\draft'^{(b)},E'^{(b)})$ \Comment{retain; other rows unchanged}
  \EndFor
\EndFor
\State \Return $\draft^{(1)},\dots,\draft^{(B)}$
\end{algorithmic}
\end{algorithm}

\subsection{Best-So-Far Invariant}
\label{app:best-so-far}

\begin{proposition}
\label{prop:app-best-so-far}
Fix the predictor, condition, and read level. Let $E'_r$ be the energy of the candidate read in round $r$, and $E_r$ the cached incumbent energy after round $r$, with $E_0=+\infty$. If $E'_1<\infty$, then for every $r\geq1$,
\[
E_r=\min_{1\leq k\leq r}E'_k,
\]
$E_r$ is non-increasing in $r$, and the incumbent after round $r$ is the earliest candidate attaining the minimum. Because a read is a deterministic function of its candidate, $E_r$ is the incumbent's energy, which therefore need not be re-read.
\end{proposition}
\begin{proof}
By induction. Round one replaces the sentinel, since $E'_1<\infty=E_0$. If $E_r=\min_{k\leq r}E'_k$, round $r+1$ either accepts, when $E'_{r+1}<E_r$, so that $E_{r+1}=E'_{r+1}=\min_{k\leq r+1}E'_k$, or keeps $E_{r+1}=E_r\leq E'_{r+1}$. Ties are rejected, so the earliest minimizer is kept.
\end{proof}

Equation~\eqref{eq:best-so-far-energy} is this proposition. It concerns completed, scored candidates only: re-noised canvases and intermediate proposal drafts are not scored, and the energy need not decrease across proposal passes. It does not assert that the incumbent minimizes the energy globally, that it is correct, or that decoding may stop once the energy stops decreasing, since the decoder runs for the fixed budget $R$. A candidate holding a token outside the canvas alphabet receives $E'=+\infty$ and is never retained.

\subsection{Model Evaluations and Cached State}
\label{app:decoding-cost}

Each round performs $J-1$ proposal passes and one read, so $R$ rounds use $R(J-1)+R=RJ$ model evaluations per condition. The placeholder is never scored, and the incumbent is never re-read. After every completed round, the incumbent is a complete output. Between rounds, the decoder carries only each condition's incumbent tokens and cached energy, and within a batch each condition accepts or rejects its candidate independently. Curves that score intermediate states, such as accuracy after each pass, are analyses rather than part of the decoder.

\subsection{Decoding Hyperparameters}
\label{app:decoding-hparams}

\begin{table}[h]

\centering
\small
\caption{Decoding configuration at the default budget of $RJ=128$ model evaluations per problem.}
\vspace{-5pt}
\label{tab:app-decode-hparams}
\resizebox{\linewidth}{!}{%
\begin{tabular}{@{}lccc@{}}
\toprule
& Countdown (CD-3/4/5) & Sudoku-Extreme & Maze-Hard\\
\midrule
Canvas alphabet $N$ / canvas length $L$ & $18$ / $64$ & $9$ / $81$ & $5$ / $900$\\
Rounds $\times$ passes per round, $R\times J$ & $32\times4$ & $32\times4$ & $16\times8$\\
Read level $\alpha_s$ & $7/8$ & $7/8$ & $7/8$\\
Re-noising keep probability $\rho_e$ & $0.1$ & $0.1$ & $0.7$\\
Positions changed by re-noising \eqref{eq:app-renoise-change} & $85.0\%$ & $80.0\%$ & $24.0\%$\\
Inner schedule & \multicolumn{3}{c}{\eqref{eq:app-inner-schedule}, with $\rho=0$ in round $1$ and $\rho=\rho_e$ afterwards}\\
Acceptance & \multicolumn{3}{c}{strictly lower cavity energy, no tolerance}\\
Decode seeds & \multicolumn{3}{c}{$3$ (random generator seeds $1001$--$1003$)}\\
Precision & \multicolumn{3}{c}{bf16 forward passes; fp32 log-softmax and energies}\\
\bottomrule
\end{tabular}}
\end{table}

Appendix~\ref{app:diag-sweep} reports the sensitivity to each.

\section{Implementation and Evaluation Details}
\label{app:implementation}

\subsection{Tasks and Data}
\label{app:tasks}

\begin{table}[h]
\centering
\small
\caption{Tasks. Lengths are in tokens; $N$ is the canvas alphabet used for corruption and scoring.}
\vspace{-5pt}
\label{tab:app-tasks}
\begin{tabular}{@{}lccc@{}}
\toprule
& Countdown (CD-3/4/5) & Sudoku-Extreme & Maze-Hard\\
\midrule
Training examples & $500$k per setting & $3{,}831{,}994$ & $1{,}000$ ($\times8$ symmetries)\\
Test examples & $1{,}000$ per setting & $422{,}786$ & $1{,}000$\\
Condition / canvas length & $20$ / $64$ & $81$ / $81$ & $900$ / $900$\\
Model vocabulary / $N$ & $31$ / $18$ & $11$ / $9$ & $6$ / $5$\\
\bottomrule
\end{tabular}
\end{table}

\paragraph{Countdown.} We use the Countdown splits released with MGDM \citep{ye2024beyond}, whose generator follows \citet{gandhi2024stream}. Inputs are $k\in\{3,4,5\}$ source numbers in $1$--$99$ and a target in $10$--$100$ and outputs are up to $k-1$ equations. The test set is target out-of-distribution: nine target values never occur in training. A solution is correct if every equation is arithmetically valid with exact division, each operand is an unused input number or an earlier result, and the final result equals the target.

\paragraph{Sudoku-Extreme.} We use the released split of \citet{wang2025hierarchicalreasoningmodel}: $3{,}831{,}994$ training puzzles, without augmentation, and $422{,}786$ test puzzles. A prediction is correct if all $81$ cells match the unique solution.

\paragraph{Maze-Hard.} We use the $1{,}000$ training and $1{,}000$ test $30\times30$ mazes of \citet{wang2025hierarchicalreasoningmodel}, augmenting training with the eight symmetries of the square. The canvas copies the maze and marks path cells. A prediction is correct if it preserves all walls and endpoints and its marked cells, together with the endpoints, form a single simple 4-connected path of shortest length, verified by breadth-first search, so any shortest path is considered correct.

\subsection{Model and Training}
\label{app:train-config}

\begin{table}[h]
\centering
\small
\caption{DRLM training configuration, shared across tasks unless noted.}
\vspace{-5pt}
\label{tab:app-train-config}
\begin{tabular}{@{}lp{0.74\linewidth}@{}}
\toprule
Architecture & bidirectional pre-LN Transformer: width $192$, $14$ layers, $4$ heads, MLP width $768$, GELU\\
 & no biases in attention and MLP projections; untied output head\\
Embeddings & shared token embedding; learned position embeddings for condition and canvas;\\
 & level features (Appendix~\ref{app:training-conventions}) added to canvas embeddings\\
Parameters & $6.27$M (Countdown), $6.28$M (Sudoku-Extreme), $6.59$M (Maze-Hard)\\
Objective & all-position cross-entropy \eqref{eq:loss}, one $\alpha\sim U(0,1)$ per sequence\\
Optimizer & AdamW, learning rate $10^{-3}$, default betas, weight decay $0.01$, gradient-norm clipping $1.0$\\
Learning-rate schedule & $200$-step linear warm-up, cosine decay to $0.1\times$ the peak\\
Batch & $256$ sequences (Maze-Hard: $64$ with $4$ accumulation steps)\\
Budget & $300$M sequences, $1{,}171{,}875$ steps (Maze-Hard: $50$M, $195{,}312$ steps)\\
Precision & bf16 autocast, TF32 matrix multiplies; dropout $0$\\
Initialization & $\mathcal N(0,0.02^2)$; residual output projections $\mathcal N(0,0.02^2/28)$\\
Hardware & 4$\times$ NVIDIA RTX A6000 \\
\bottomrule
\end{tabular}
\end{table}

\subsection{Baselines}
\label{app:baselines}

Baselines trained here use our data splits, correctness criteria, and checkpoint rule, applied to their own decoders. 

\paragraph{Diffusion baselines.} UDLM \citep{schiff2024discreteguidance} uses our backbone (Table~\ref{tab:app-train-config}) with the same data, optimizer, and budget as DRLM. It differs only in its objective, the continuous-time uniform-noise NELBO, and its decoder, the ancestral sampler with $T=128$ denoising steps. SUNDAE \citep{savinov2021step} also uses our backbone with the same optimizer and training budget, and follows the original objective and sampler. It is trained by unrolled denoising with no corruption level conditioning: the model denoises all positions once, a second pass denoises a sample of its own first-pass prediction, and the loss averages the cross-entropy of both passes over all positions. MGDM \citep{ye2024beyond} is trained with its official code and 6M recipe (a 3-layer GPT-2 of width $384$) with identical training budget as DRLM, and decoded with its top-$k$ decoder for $T=20$ steps. In Countdown, numbers differ from \citet{ye2024beyond} due to the differences in evaluation. We consider solutions invalid if the numbers aren't provided in the input or derived by an earlier result, following \citet{jo2026loopholing}. LDDM-G \citep{jo2026loopholing} adds loopholing to the MGDM recipe (self-conditioning rate $0.9$) and decodes for $T=20$ steps, keeping unmasked tokens fixed.

\paragraph{Autoregressive baselines.} Countdown results for GPT-2 trained from scratch and for LLaMA-7B/13B are taken from \citet{ye2024beyond}. On Sudoku-Extreme and Maze-Hard, we fine-tune LLaMA-13B with the LoRA \citep{hu2022lora} recipe in the official code of \citet{ye2024beyond} (rank $16$, learning rate $10^{-4}$, batch $128$) for its number of optimizer steps. 

\paragraph{Recurrent models.} HRM\citep{wang2025hierarchicalreasoningmodel} and TRM\citep{jolicoeurmartineau2025morerecursivereasoningtiny} reported results are from \citet{jolicoeurmartineau2025morerecursivereasoningtiny}. Reported numbers on Sudoku-Extreme are trained and evaluated on a small training set (1K) with $1,000$ shuffling augmentation per data example. Their Maze-Hard metric requires an exact match to a single reference path, whereas we accept any shortest path, verified by breadth-first search, since every test maze admits multiple shortest paths.

\section{Additional Experiments}
\label{app:diagnostics}

All diagnostics use the default configuration of Table~\ref{tab:app-decode-hparams} and change one component at a time. Entries are Pass@1 (\%) at $128$ model evaluations, averaged over three decode seeds ($\pm$: standard deviation over seeds). The tables are generated from the run logs.

\subsection{Effect of the Self-Likelihood Correction}
\label{app:diag-cavity-vs-raw}

To isolate the self-likelihood correction, we replace the cavity energy with the raw self-token energy $E^{\rm raw}_\theta(\draft)=-\sum_i\log d_i$, computed from the same read, and keep everything else fixed: the model, the read level, the proposal schedule, and the budget. At a fixed read level, both scores sum one term per position, and both terms decrease in the read probability $d_i$: the raw term is $-\log d_i$, and by \eqref{eq:app-self-cavity} the cavity term is $\log\big(1+\kappa_{\alpha_s}(1-d_i)/d_i\big)$. Since $\kappa_{\alpha_s}$ grows with the read level, the decomposition predicts that the correction matters more as $\alpha_s$ increases.

We test this prediction at five read levels. At each level we run the full decoder once with each score, with the proposal schedule ending at that level as in Algorithm~\ref{alg:app-refinement-batched}, and pair the two runs over problems. 

\begin{table}[h]
\centering
\small
\caption{Self-likelihood correction on Sudoku-Extreme. Pass@1 at $128$ evaluations when the decoder accepts candidates by the cavity or by the raw energy, both read at level $\alpha_s$.}
\vspace{-5pt}
\label{tab:app-diag-levels}
\begin{tabular}{@{}lccccc@{}}
\toprule
$\alpha_s$ & $\kappa_{\alpha_s}$ & Cavity & Raw & Difference (Cavity $-$ Raw)\\
\midrule
$3/4$ & $28$ & $94.0$ & $88.8$ & $+5.2$ \\
$7/8$ & $64$ & $93.9$ & $88.7$ & $+5.2$ \\
$15/16$ & $136$ & $94.0$ & $88.6$ & $+5.4$ \\
$0.98$ & $442$ & $94.3$ & $88.5$ & $+5.8$ \\
$0.995$ & $1{,}792$ & $94.3$ & $88.5$ & $+5.8$ \\
\bottomrule
\end{tabular}
\end{table}
Table~\ref{tab:app-diag-levels} compares decoding accuracy on Sudoku-Extreme under cavity and raw energy scoring. The cavity energy leads at every read level, and the lead widens with $\kappa_{\alpha_s}$: $+5.2$ points at the default $\alpha_s=7/8$ ($\kappa_{\alpha_s}=64$), $+5.4$ at $15/16$, and $+5.8$ at both $0.98$ and $0.995$ ($\kappa_{\alpha_s}\ge 442$), where the paired intervals exclude zero. The two decoders drift apart as the read level rises: the cavity decoder improves from $93.9\%$ to $94.3\%$, while the raw decoder slips from $88.8\%$ to $88.5\%$. These results show that correcting for self-token evidence improves draft refinement over raw self-token scoring at the same inference budget. The larger gains at higher read levels are consistent with the increasing self-likelihood advantage supplied by the corruption channel.

\subsection{Sensitivity at a Fixed Budget}
\label{app:diag-sweep}

Table~\ref{tab:app-diag-sweep} varies one hyperparameter at a time with $RJ=128$ fixed, so $R=128/J$. Bold entries are the defaults.

\begin{table}[h]
\centering
\small
\caption{Pass@1 at $RJ=128$ when one hyperparameter is changed.}
\label{tab:app-diag-sweep}
\begin{tabular}{@{}lccc@{}}
\toprule
Setting & CD-4 & CD-5 & Maze-Hard\\
\midrule
$J=2$ & $82.7$ & $26.3$ & $67.5$\\
$J=4$ & \textbf{$87.6$} & \textbf{$44.4$} & $72.3$\\
$J=8$ & $86.9$ & $41.7$ & \textbf{$73.0$}\\
$J=16$ & $86.2$ & $36.7$ & $72.4$\\
$\alpha_s=3/4$ & $87.7$ & $44.9$ & $71.5$\\
$\alpha_s=7/8$ & \textbf{$87.6$} & \textbf{$44.4$} & \textbf{$73.0$}\\
$\alpha_s=15/16$ & $87.7$ & $44.7$ & $73.4$\\
$\rho_e=0$ (restart) & $85.3$ & $30.6$ & $66.2$\\
$\rho_e=0.1$ & \textbf{$87.6$} & \textbf{$44.4$} & --\\
$\rho_e=0.3$ & $87.0$ & $35.5$ & --\\
$\rho_e=0.5$ & -- & -- & $72.2$\\
$\rho_e=0.7$ & -- & -- & \textbf{$73.0$}\\
\bottomrule
\end{tabular}
\end{table}

One proposal pass per round ($J=2$) is the worst setting on every task, and the defaults are the best tested values of $J$. Between $\alpha_s=3/4$ and $15/16$ the read level changes Countdown accuracy by at most $0.5$ points and Maze-Hard accuracy by $1.9$ points. Restarting from noise ($\rho_e=0$) is the worst re-noising setting on every task, and on CD-5 keeping more of the incumbent ($\rho_e=0.3$) is also clearly worse than the default.

For Sudoku-Extreme we selected the configuration on the $2{,}000$-puzzle subsample among four candidates (Table~\ref{tab:app-sudoku-selection}) and fixed it before evaluating the remaining $420{,}786$ puzzles. Restarting from noise ($\rho_e=0$, $J=4$), which was not a candidate, reaches $77.0\pm0.2$ on the subsample: above keeping most of the incumbent ($\rho_e=0.7$), but far below the selected $\rho_e=0.1$.

\begin{table}[h]
\centering
\small
\caption{Sudoku-Extreme: configuration selection on the $2{,}000$-puzzle subsample, and the selected configuration on the held-out complement and the full test set.}
\label{tab:app-sudoku-selection}
\begin{tabular}{@{}lc@{}}
\toprule
Configuration & Pass@1\\
\midrule
Subsample: $J=4$, $\rho_e=0.1$ & \textbf{$93.9\pm0.2$}\\
Subsample: $J=4$, $\rho_e=0.7$ & $74.5\pm0.1$\\
Subsample: $J=8$, $\rho_e=0.1$ & $91.3\pm0.4$\\
Subsample: $J=8$, $\rho_e=0.7$ & $75.1\pm0.3$\\
\midrule
Selected, held-out complement ($420{,}786$ puzzles, 3 seeds) & $93.9\pm0.0$\\
Selected, full test set ($422{,}786$ puzzles) & $93.9\pm0.0$\\
\bottomrule
\end{tabular}
\end{table}

\subsection{Budget, Acceptance, and Coverage}
\label{app:diag-budget}

Table~\ref{tab:app-diag-budget} reports Pass@1 as the budget grows and the fraction of problems whose incumbent is replaced in a given round. \emph{Coverage} is the fraction of problems for which at least one completed candidate of the run was correct which upper-bounds what any rule selecting among these candidates could achieve.

\begin{table}[h]
\centering
\small
\caption{Pass@1 by model evaluations, coverage at $128$ evaluations, and the fraction of problems whose incumbent is replaced in the last round. }
\label{tab:app-diag-budget}
\resizebox{\linewidth}{!}{%
\begin{tabular}{@{}lcccccccc@{}}
\toprule
& \multicolumn{6}{c}{Pass@1 after this many model evaluations} & & \\
\cmidrule(lr){2-7}
Task & $4$ & $8$ & $16$ & $32$ & $64$ & $128$ & Coverage & Accepted, last round (\%)\\
\midrule
CD-4 & $82.8$ & $83.8$ & $84.8$ & $85.9$ & $86.6$ & $87.6$ & $88.6$ & $0.10$\\
CD-5 & $24.6$ & $26.5$ & $29.3$ & $33.5$ & $38.2$ & $44.4$ & $44.8$ & $1.93$\\
Maze-Hard & -- & $62.9$ & $69.3$ & $70.7$ & $72.2$ & $73.0$ & $73.9$ & $0.47$\\
\bottomrule
\end{tabular}}
\end{table}

Pass@1 stays within one point of coverage on every task, so the energy rarely discards a correct candidate that the decoder has proposed. The remaining failures are mostly problems for which no correct candidate appears within the budget. The examples in Appendix~\ref{app:qual} show both kinds of failure. On Maze-Hard, most repairs by a retained candidate remove extra path cells, while a minority reroute part of the path or fill a gap. Across the rounds of decode seed $1$ that turned an invalid draft into a solved one, the number of wrong path cells never increased from one proposal pass to the next.

\subsection{Objective and Decoder}
\label{app:diag-objective-decoder}

Table~\ref{tab:ablation} crosses the training objective with the decoder on CD-5 at $128$ model evaluations. Table~\ref{tab:app-objective-decoder} gives the cells with standard deviations over three decode seeds and the paired effects with $95\%$ bootstrap intervals over problems. The UDLM model has the backbone of Table~\ref{tab:app-train-config} and is trained with UDLM's NELBO on the same data and budget. Its predictions target the leave-one-out posterior, so under our decoder they are converted to full posteriors by adding $\log\kappa_\alpha$ to the observed token's logit over the canvas alphabet (Proposition~\ref{prop:cavity-decomposition}), for proposals and reads alike. Under the ancestral sampler ($T=128$, the kernel of Proposition~\ref{prop:app-uniform-mixture}), DRLM's full posteriors are converted to leave-one-out predictions by subtracting $\log\kappa_{\alpha_t}$ at every step. The conversions change the cross cells by at most $1.3$ points (rows ``without conversion'').

\begin{table}[h]
\centering
\small
\caption{Objective $\times$ decoder on CD-5 at $128$ model evaluations: Pass@1 per cell ($\pm$: standard deviation over decode seeds) and paired effects (points) with $95\%$ bootstrap intervals over problems.}
\label{tab:app-objective-decoder}
\begin{tabular}{@{}lc@{}}
\toprule
Cell or effect & Pass@1 or difference\\
\midrule
DRLM (CE) $\times$ refinement decoder & $44.4\pm1.0$\\
UDLM (NELBO) $\times$ refinement decoder & $36.6\pm0.6$\\
DRLM (CE) $\times$ ancestral sampler & $22.5\pm0.3$\\
UDLM (NELBO) $\times$ ancestral sampler & $6.9\pm0.5$\\
UDLM $\times$ refinement decoder, without conversion & $37.3\pm0.8$\\
DRLM $\times$ ancestral sampler, without conversion & $21.2\pm0.3$\\
\midrule
Objective, under the refinement decoder & $+7.8$ $[+5.7,+9.9]$\\
Objective, under the ancestral sampler & $+15.6$ $[+13.6,+17.7]$\\
Decoder, DRLM model & $+21.9$ $[+19.8,+24.1]$\\
Decoder, UDLM model & $+29.7$ $[+27.5,+32.1]$\\
Interaction & $-7.8$ $[-10.5,-5.2]$\\
\bottomrule
\end{tabular}
\end{table}

Both the objective and the decoder improve accuracy under either choice of the other. The effects are not additive: the refinement decoder helps the UDLM model more than the DRLM model, and the objective matters more under the ancestral sampler.

\section{Qualitative Examples}
\label{app:qual}

\begin{figure}[h]
\centering
\includegraphics[width=\textwidth]{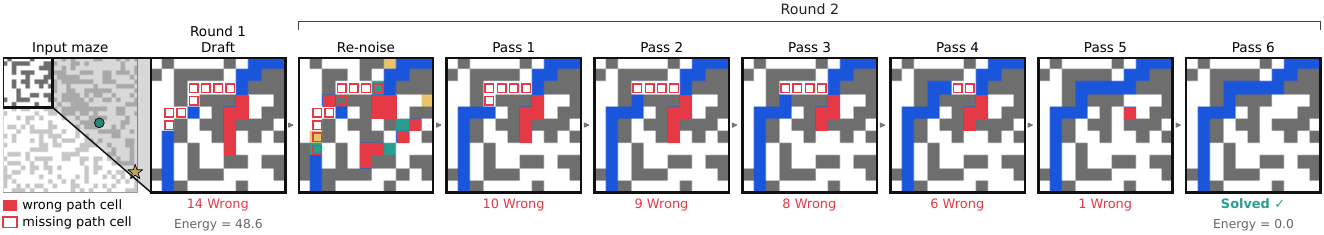}\\[-2pt]
{\small (a) The first draft follows a wrong corridor; one round refines the path onto the correct one over six passes.}\\[6pt]
\includegraphics[width=\textwidth]{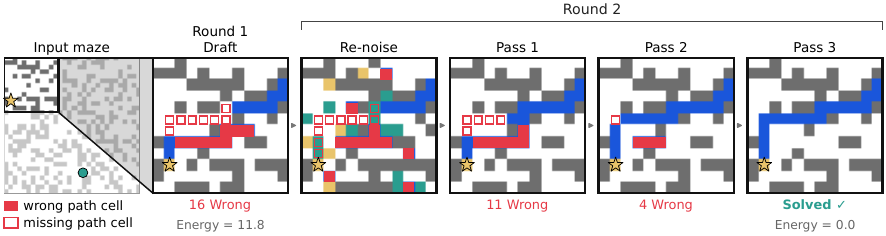}\\[-2pt]
{\small (b) Near the goal, the first draft uses the row below the shortest path; one round refines it.}\\[6pt]
\includegraphics[width=0.85\textwidth]{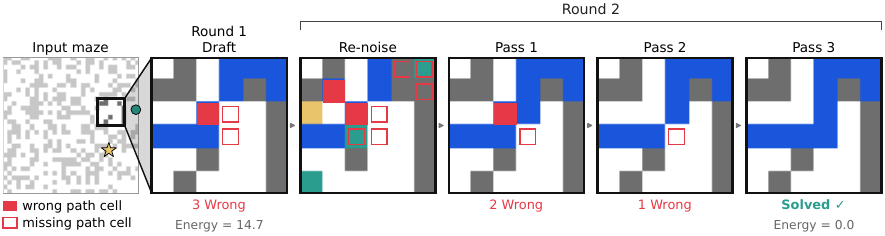}\\[-2pt]
{\small (c) The first draft is broken; one round connects the two segments.}
\caption{Maze-Hard repairs by a single retained proposal. Panels after the input show the boxed region. Red fill marks path cells not in the final path, and red outlines mark final-path cells not yet drawn.}
\label{fig:app-maze-gallery}
\end{figure}

\begin{figure}[h]
\centering
\includegraphics[width=\textwidth]{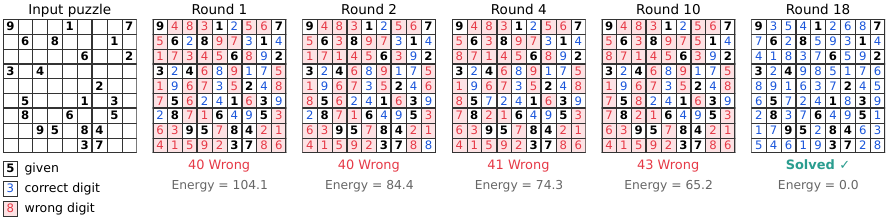}\\[-2pt]
{\small (a) The drafts retained for one puzzle, each with lower energy than the one it replaces, until round~$18$ retains the solution.}\\[6pt]
\includegraphics[width=0.663\textwidth]{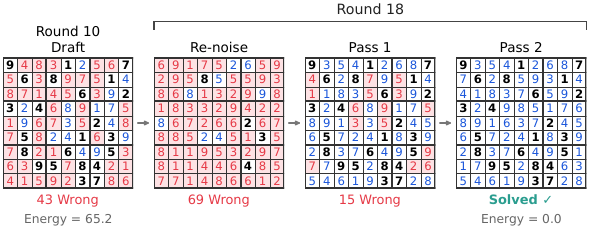}\\[-2pt]
{\small (b) Round~$18$ of the same puzzle: re-noising replaces most cells of the round-$10$ draft, and two passes reach the solution.}\\[6pt]
\includegraphics[width=0.497\textwidth]{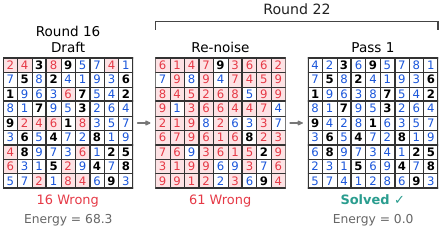}\\[-2pt]
{\small (c) Another puzzle: one pass solves it from the re-noised round-$16$ draft.}
\caption{Sudoku-Extreme refinement on two puzzles of the $2{,}000$-puzzle subsample. Bold digits are givens, blue digits match the unique solution, and red cells differ from it. Re-noising acts on all $81$ cells, givens included. Each round has three proposal passes before its read; passes after the first solved one are omitted.}
\label{fig:app-sudoku}
\end{figure}

All examples come from the reported runs (default configuration). Figure~\ref{fig:app-maze-gallery} shows Maze-Hard rounds in the format of Figure~\ref{fig:qual_mze}, Table~\ref{tab:app-cd-examples} shows Countdown-4 rounds as text, and Figure~\ref{fig:app-sudoku} shows Sudoku-Extreme puzzles from the $2{,}000$-puzzle subsample.

On every task, a round re-noises the retained draft, rebuilds a complete candidate over its passes, and keeps the candidate only if its energy is lower. The tasks differ in how much of the draft re-noising keeps, and hence in how much a successful revision changes. On Maze-Hard ($\rho_e=0.7$), a revision that turns an invalid draft into a valid one changes a median of $2.5$ of the $900$ cells. On Countdown and Sudoku-Extreme ($\rho_e=0.1$), it rewrites a large part of the answer: a median of $13$ of $26$ characters on CD-4, $14$ of $33$ on CD-5, and $28$ of $81$ cells on Sudoku-Extreme.).

\begin{table}[h]
\centering
\small
\caption{Countdown-4 rounds. Each candidate is read at $\alpha_s=7/8$ and kept only if its energy is lower than the incumbent's. Rounds whose candidate repeats the incumbent are omitted.}
\label{tab:app-cd-examples}
\begin{tabular}{@{}clccc@{}}
\toprule
Round & Candidate & Valid & Energy & Kept\\
\midrule
\multicolumn{5}{@{}l}{\emph{Numbers $9,7,13,21$; target $89$: the plan is replaced}}\\
1 & \texttt{21-7=12,9*13=102,102-13=89} & no & $15.73$ & yes\\
2 & \texttt{9*13=117,7+21=28,117-28=89} & yes & $0.06$ & yes\\
\midrule
\multicolumn{5}{@{}l}{\emph{Numbers $81,4,1,64$; target $96$: one token is repaired, and a worse candidate is rejected}}\\
1 & \texttt{81-1=80,64/4=14,80+16=96} & no & $18.72$ & yes\\
2 & \texttt{81-1=80,64/4=16,80+16=96} & yes & $0.00$ & yes\\
3 & \texttt{81+1=80,64/4=14,80+14=96} & no & $15.83$ & no\\
\midrule
\multicolumn{5}{@{}l}{\emph{Numbers $42,47,9,14$; target $81$: failure, a lower-energy invalid candidate replaces a valid draft}}\\
1 & \texttt{47-42=5,14-5=9,9*9=81} & yes & $2.40$ & yes\\
9 & \texttt{47-42=5,14+5=9,9*9=81} & no & $2.00$ & yes\\
\bottomrule
\end{tabular}
\end{table}

\begin{figure}[h]
\centering
\includegraphics[width=0.5\textwidth]{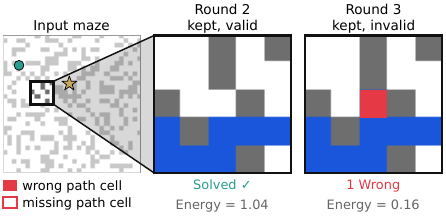}
\caption{A Maze-Hard failure case. Round~$2$ retains a valid shortest path while round~$3$ proposes the same path with one extra cell, whose lower energy leads the decoder to replace the valid draft.}
\label{fig:app-maze-failure}
\end{figure}

Single-token repairs such as the second example are the exception on Countdown: only $17\%$ of the revisions that fix an invalid CD-4 draft change at most three characters ($4\%$ on CD-5). The failure in Table~\ref{tab:app-cd-examples} illustrates the limit of a learned score: the energy measures the model's estimated contextual support for each token, not arithmetic validity, and here it prefers an invalid draft. Figure~\ref{fig:app-maze-failure} shows the same failure on Maze-Hard, where a one-cell spur on a solved path lowers the energy and is retained. Such ranking errors are rare in aggregate (Appendix~\ref{app:diag-budget}).

\paragraph{Sudoku-Extreme.} Figure~\ref{fig:app-sudoku} shows the same round structure on Sudoku. As on Maze-Hard, the round that fixes a draft removes errors progressively over its passes: for all $527$ subsample puzzles that are wrong after round~$1$ and solved within $128$ evaluations (decode seed~$1$), the number of wrong cells never increases from one pass of the solving round to the next. A solving revision changes more cells than on Maze-Hard because the errors of a retained draft are coupled. A median of only $20\%$ of its wrong cells conflict with another cell of their row, column, or box, and a median of $70\%$ of the rows, columns, and boxes that contain an error still hold all nine digits. The wrong cells of such a unit permute its correct digits, so they can only be corrected together. Before the solving round, retained drafts replace one another with decreasing energy while staying about as far from the solution (Figure~\ref{fig:app-sudoku}a). 

\end{document}